\pdfoutput=1

\documentclass[11pt]{article}

\usepackage[]{acl}

\usepackage{times}
\usepackage{latexsym}

\usepackage[T1]{fontenc}

\usepackage[utf8]{inputenc}

\usepackage{color}
\usepackage[table]{xcolor}
\usepackage{tcolorbox}
\usepackage{xcolor}
\usepackage{tabularx}
\usepackage{tabu}
\usepackage{booktabs}
\usepackage{multirow}
\usepackage{makecell}

\usepackage{graphicx} 
\usepackage{float}
\usepackage{subfigure} 
\usepackage{amssymb}

\usepackage{hyperref}
\usepackage{tablefootnote}
\usepackage{xspace}

\usepackage{float}
\usepackage{amsmath}
\usepackage{bm}
\usepackage{algorithmicx}
\usepackage{algorithm}
\usepackage{algpseudocode}

\usepackage{arydshln}

\usepackage{amssymb}
\usepackage{pifont}
\usepackage{enumitem}
\usepackage{fontawesome5}
\usepackage{microtype}

\newcommand{\ours}{Table-R1\xspace}
\newcommand{\sftmodel}{\ours-SFT\xspace}
\newcommand{\rlmodel}{\ours-Zero\xspace}
\newcommand{\eg}{\hbox{\emph{e.g.,}}\xspace}
\newcommand{\ie}{\hbox{\emph{i.e.,}}\xspace}
\newcommand{\huggingface}{\raisebox{-1.5pt}{\includegraphics[height=1.05em]{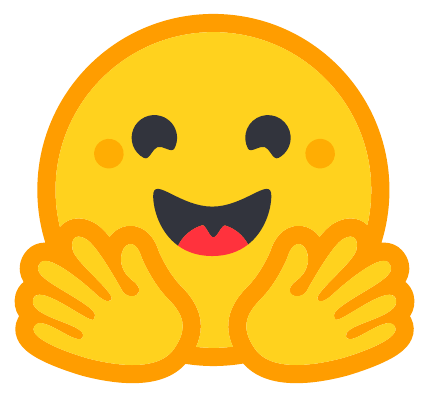}}\xspace}
\newcommand{\github}{\raisebox{-1.5pt}{\includegraphics[height=1.05em]{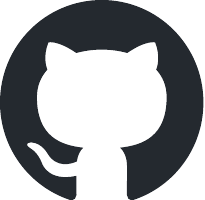}}\xspace}

\newcommand{\cdashlinelr}[1]{
  \cdashline{#1}[.4pt/2pt]
}

\addtocontents{toc}{\protect\setcounter{tocdepth}{-1}}
\title{\ours: Inference-Time Scaling for Table Reasoning}

\author{
Zheyuan Yang\thanks{Equal Contributions.}$^1$ \quad Lyuhao Chen$^{*2}$ \quad Arman Cohan$^1$ \quad Yilun Zhao$^1$ \vspace{10pt}\\
$^{1}$Yale NLP Lab \quad $^{2}$Carnegie Mellon University
}

\begin{document}
\maketitle

\begin{abstract}
In this work, we present the first study to explore inference-time scaling on table reasoning tasks. We develop and evaluate two post-training strategies to enable inference-time scaling: distillation from frontier model reasoning traces and reinforcement learning with verifiable rewards (RLVR). 
For distillation, we introduce a large-scale dataset of reasoning traces generated by DeepSeek-R1, which we use to fine-tune LLMs into the \sftmodel model. 
For RLVR, we propose task-specific verifiable reward functions and apply the GRPO algorithm to obtain the \rlmodel model. 
We evaluate our \ours-series models across diverse table reasoning tasks, including short-form QA, fact verification, and free-form QA. 
Notably, the \rlmodel model matches or exceeds the performance of GPT-4.1 and DeepSeek-R1, while using only a 7B-parameter LLM. It also demonstrates strong generalization to out-of-domain datasets.
Extensive ablation and qualitative analyses reveal the benefits of instruction tuning, model architecture choices, and cross-task generalization, as well as emergence of essential table reasoning skills during RL training.

\vspace{-10pt}
\begin{small}
\begin{center}
\begin{tabular}{ll}
\huggingface~~\textbf{Model} &  \href{https://huggingface.co/Table-R1} {\path{huggingface.co/Table-R1}}\\
\github~~\textbf{Code} &\href{https://github.com/Table-R1/Table-R1}{\path{github.com/Table-R1}}\\
\end{tabular}
\end{center} 
\end{small}

\end{abstract}

\section{Introduction}
Reasoning large language models, such as OpenAI’s o-series~\cite{jaech2024openai,pfister2025understanding} and Deepseek’s R1~\cite{guo2025deepseek}, have demonstrated enhanced reasoning capabilities by inference-time scaling, \ie generating a reasoning chain of tokens that allow the model to ``think'' before giving the final answer.
Building on this success, recent research has extended inference-time scaling to various domains and tasks, including multimodal reasoning~\cite{huang2025visionr1, xu2025llavacot}, machine translation~\cite{feng2025mt}, agent-based tool use~\cite{Agent-R1, jin2025searchr1}, and information retrieval~\cite{weller2025rank1, zhuang2025rankr1}. 

\begin{figure}[!t]
  \includegraphics[width=\columnwidth]{figure/RadarChart.pdf}
  \caption{Overall performance comparison between \ours and same-scale baselines on various table reasoning benchmarks. Both \sftmodel and \rlmodel exhibit substantial performance improvements over baselines, showing the effectiveness of our approach across both in- and out-of-domain benchmarks.}
  \label{fig:experiments}
\end{figure}

However, applying inference-time scaling to structure-dependent tasks—particularly table reasoning—remains largely unexplored. 
Table reasoning presents distinct challenges compared to text-only tasks: it requires interpreting diverse cell contents, aligning data across the table, and performing multi-step reasoning with aggregation and numerical operations~\cite{deng2024tables, wu2025tablebench}. These requirements are further complicated by the need to process long and densely structured tabular inputs~\cite{zhao-etal-2023-investigating, nahid2024tabsqlify, zhang2025survey}.
Advancing LLMs’ reasoning capabilities over tabular tasks holds significant promise for real-world applications, including data analysis~\cite{zhao-etal-2024-docmath}, scientific reporting~\cite{liang2024mapping, newman2024arxivdigestables}, and decision-support systems~\cite{handler2024large}.

\begin{figure}[!t]
  \includegraphics[width=\columnwidth]{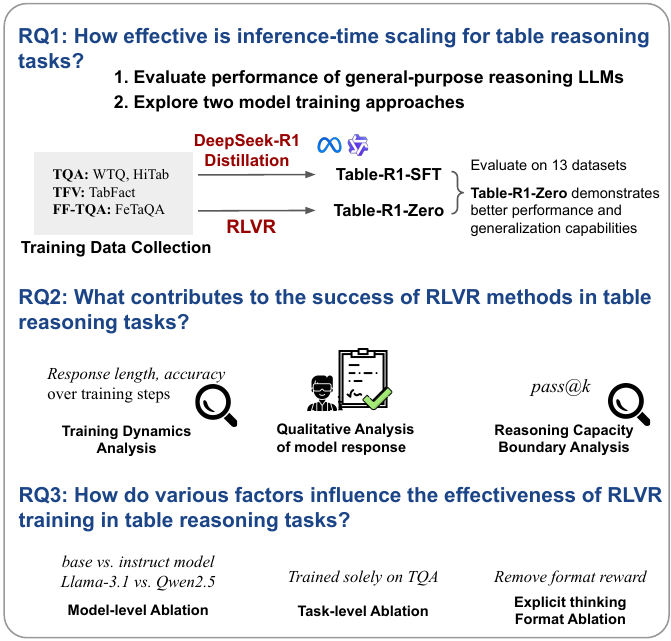}
  \caption{An overview of our research and three research questions investigated in this study.}
  \label{fig:overview}
\end{figure}

In this work, we present the first study to explore inference-time scaling on table reasoning tasks. 
\autoref{fig:overview} presents the overview of our research.
We develop and systematically evaluate two widely used post-training strategies to enable inference-time scaling on table reasoning tasks: (1) distilling from reasoning traces of frontier reasoning models, and (2) reinforcement learning with verifiable rewards (RLVR).
For the distillation approach, we curate and open-source a large-scale table reasoning dataset containing reasoning traces generated by DeepSeek-R1 and verifid by LLM-based annotators. We fine-tune LLMs on this data to obtain \sftmodel.
For the RLVR approach, we design task-specific, verifiable reward functions tailored to table reasoning and apply the Group Relative Policy Optimization (GRPO) algorithm~\cite{shao2024deepseekmath,guo2025deepseek} to enable stable and scalable reinforcement learning. This yields the \rlmodel model.

We evaluate \ours-series models on a wide range of table reasoning tasks, including short-form table QA, fact verification, and free-form table QA.
Our experiments demonstrate the effectiveness of inference-time scaling for table reasoning.
The RLVR approach, in particular, exhibits better performance and generalization capabilities, compared to the distillation approach. 
Notably, our \rlmodel models achieve performance that is competitive with advanced language models such as GPT 4.1 and DeepSeek R1, despite using only a 7B-parameter LLM (\ie Qwen2.5-7B) as the backbone.
We further conduct comprehensive ablation studies on \emph{instruction tuning benefits}, \emph{model family comparisons}, and \emph{cross-task generalization}, providing insights for future applications of inference-time scaling in table reasoning.
Our qualitative analysis of model responses reveals that \rlmodel not only acquires multi-step reasoning and reflection abilities like other reasoning models, but also develops essential table-specific reasoning skills such as semantic understanding, information extraction, and arithmetic computation.

\section{Related Work}
\subsection{Inference-Time Scaling}
Recently, OpenAI's o1 has demonstrated that scaling inference-time computation can significantly enhance the reasoning abilities of large language models (LLMs) on complex tasks~\cite{jaech2024openai}. To leverage this, various inference-time strategies have been explored, including the use of Monte Carlo Tree Search (MCTS) for exploring diverse reasoning trajectories~\cite{feng2023alphazero, qi2024mutual, guan2025rstar} and process reward models (PRMs) that offer step-level feedback to guide model outputs~\cite{lightman2023let, yuan2024free}. In parallel, supervised fine-tuning (SFT) on reasoning traces has emerged as a practical post-training method, enabling LLMs to better align generation with explicit chain-of-thought reasoning patterns~\cite{wen2025light, muennighoff2025s1, ye2025limo}.

Beyond supervised approaches, recent work has introduced reinforcement learning from verifiable rewards (RLVR) as a promising post-training paradigm for LLM reasoning~\cite{guo2025deepseek, team2025kimi, qwq32b}. In this setting, models are directly optimized with rule-based rewards, allowing them to autonomously discover effective reasoning strategies without explicit intermediate supervision. Subsequent studies have improved RLVR training by incorporating dynamic sampling, token-level policy gradients, and reward normalization to enhance training stability and sample efficiency~\cite{yu2025dapo, liu2025understanding, xia2025mimo}. The RLVR paradigm has demonstrated strong generalization across diverse domains, including mathematical problem solving~\cite{hu2025open, openr1}, logical reasoning games~\cite{xie2025logic}, vision-based reasonings~\cite{huang2025vision}, and interactive agent scenarios~\cite{wang2025ragen, xia2025gui, feng2025retool}.

\subsection{Table Reasoning}
Reasoning over tabular data has long attracted attention due to its practical applications in real-world scenarios such as data analysis. It encompasses a variety of tasks, including short-form question answering~\cite{pasupat-liang-2015-compositional, cheng-etal-2022-hitab, lu2023dynamic, zhao-etal-2023-robut, zhao-etal-2024-knowledgefmath, wu2025tablebench}, fact verification~\cite{Chen2020TabFact, gupta-etal-2020-infotabs}, and free-form question answering~\cite{nan-etal-2022-fetaqa, zhao-etal-2023-loft}.
Earlier research primarily focused on fine-tuning smaller language models for specific tasks~\cite{herzig-etal-2020-tapas, zhao-etal-2022-reastap, liu2022tapex}. 
Recent efforts have advanced the adaptation of LLMs for table reasoning~\cite{zhang-etal-2024-tablellama, zhang2025tablellm, su2024tablegpt2, zha2023tablegpt, su2024tablegpt2, deng2025betterunderstandingtableinstruction}, enabling more general-purpose capabilities across task types. 
In parallel, several studies have begun exploring agentic approaches to tackle table reasoning tasks~\cite{10.1145/3539618.3591708, zhao-etal-2024-tapera, nan-etal-2024-evaluating, yu2025tablecriticmultiagentframeworkcollaborative}.
Despite these advances, the application and enhancement of inference-time scaling for table reasoning remain largely underexplored. 
Our study shows that 7B-scale LLMs with inference-time scaling can match the performance of frontier models such as GPT-4.1. This finding highlights a promising direction for advancing table reasoning.

\section{\ours Models}
To systematically explore inference scaling in table reasoning tasks, we develop two variants of \ours model, each leveraging a widely adopted post-training strategy for inference-time scaling:
(1) \sftmodel, trained via supervised fine-tuning on reasoning traces generated by frontier reasoning LLMs, and
and (2) \rlmodel, trained using our developed RLVR approach tailored for table reasoning tasks.
The methodologies for each approach are detailed in the following subsections.

\subsection{Training Data Collection}
We construct the \ours training dataset by integrating three representative table reasoning tasks, each introducing distinct reasoning challenges: (1) Short-form Table QA, which requires models to provide precise answers to questions grounded in tabular data; (2) Table Fact Verification, which requires models to determine whether a given claim is entailed by the table content; and (3) Free-form Table QA, which requires models to produce open-ended answers grounded in tabular information.
Each task contributes unique reasoning challenges, and we select established datasets (presented in \autoref{tab:training-data}) to ensure comprehensive coverage. 
All datasets are preprocessed with the presence of verifiable ground truths for reward computation to align with the RLVR paradigm. More detailed explanation of training data is presented in Appendix~\ref{sec:appendix-train-dataset}.

\begin{table}[!t]
	\centering
	\small
    \setlength{\tabcolsep}{1pt}
    \resizebox{\linewidth}{!}{
	\begin{tabular}{llr}
		\toprule
		\textbf{Task} & \textbf{Dataset} & \textbf{Samples} \\
		\midrule
		\multirow{2}*{Short-form QA (TQA)} 
        & WTQ~\cite{pasupat-liang-2015-compositional} & 13,706 \\
        & HiTab~\cite{cheng-etal-2022-hitab} & 6,793 \\
		\midrule
	   Fact Verification (TFV) & TabFact~\cite{Chen2020TabFact} & 20,740 \\
		\midrule
		Free-form QA (FF-TQA) & FeTaQA~\cite{nan-etal-2022-fetaqa} & 7,324 \\
		\bottomrule
	\end{tabular}
	}
	\caption{Overview of datasets collected in \ours training data. For each dataset, we sample examples from its training set.}
	\label{tab:training-data}
\end{table}

\subsection{\sftmodel via Supervised Finetuning}
To enable inference-time scaling in \sftmodel, we curate a new table reasoning dataset featuring long CoT reasoning.
Specifically, for each instance in the raw training data described in the previous subsection, we use DeepSeek-R1~\cite{deepseekr1} to generate a long CoT response. The response consists of a step-by-step reasoning process followed by a final answer. We present the prompts for response generation in Appendix~\ref{sec:appendix-prompt-template}.
To ensure the quality and correctness of the training data, we apply automated evaluators (detailed in Section~\ref{sec:eval}) to assess the final answers. Examples with incorrect answers are filtered out. 
After this verification step, we obtain a total of 33,601 high-quality examples for SFT training.
This dataset is then used to train the \sftmodel model.

\subsection{\rlmodel via RL Training}
We describe the RLVR algorithm and the reward design for training \rlmodel as follows.

\paragraph{Reinforcement Learning with Verifiable Rewards (RLVR).}
We adapt Group Relative Policy Optimization (GRPO) with recent improvements introduced by DAPO~\cite{yu2025dapo}, including both token-level loss computation and asymmetric (decoupled) clipping. Notably, following recent work~\cite{hu2025open, liu2025understanding, xia2025mimo}, we omit the KL penalty term present in the original GRPO~\cite{jaech2024openai,pfister2025understanding}.
We formally define the RL algorithm applied in our study as follows: 

For each input $(q, a)$, the policy $\pi_\theta$ samples a group of $G$ candidate responses $\{o_i\}_{i=1}^G$. Each response receives a reward $R_i$ as described in the next paragraph. 
The group-normalized advantage for the $i$-th response at time step $t$ is:

\begin{small}
\begin{equation}
\hat{A}_{i,t} = \frac{R_i - \text{mean}(\{R_j\}_{j=1}^G)}{\text{std}(\{R_j\}_{j=1}^G)}.
\end{equation}
\end{small}

\noindent Our objective is optimized at the token level with decoupled, asymmetric clipping:

\begin{small}
\begin{equation}
\begin{aligned}
\mathcal{J}_{\text{GRPO}}(\theta) 
&= \mathbb{E}_{(q,a)\sim \mathcal{D},\, \{o_i\}_{i=1}^G\sim \pi_{\theta_\text{old}}(\cdot\mid q)} \\
&\Bigg[
  \frac{1}{\sum_{i=1}^{G}|o_i|}\sum_{i=1}^{G}\sum_{t=1}^{|o_i|} \min \Big( r_{i,t}(\theta) \hat{A}_{i,t}, \\
&\text{clip}\big(r_{i,t}(\theta), 1-\varepsilon_{\text{low}}, 1+\varepsilon_{\text{high}}\big)\hat{A}_{i,t} \Big)
\Bigg]
\end{aligned}
\label{eq:grpo}
\end{equation}
\end{small}

\noindent where the probability ratio $r_{i,t}(\theta)$ is defined as:

\begin{small}
\begin{equation}
r_{i,t}(\theta) = \frac{\pi_\theta(o_{i,t} \mid q, o_{i,<t})}{\pi_{\theta_{\text{old}}}(o_{i,t} \mid q, o_{i,<t})}.
\end{equation}
\end{small}

\noindent This formulation enables stable and effective RL fine-tuning for table reasoning with LLMs.

\paragraph{Reward Design.}
To facilitate effective RL training, we design verifiable reward signals tailored to the characteristics of each table reasoning task. Our reward framework consists of two components: \emph{accuracy rewards} and \emph{format rewards}.
Accuracy rewards measure the correctness of model outputs. We define task-specific reward functions as follows:
\begin{itemize} [leftmargin=*]
\itemsep0em
    \item \textbf{TQA}: The ground-truth is a short-answer list, where each element contains several words. We employ exact match to assign a reward of 1 for a correct answer and 0 otherwise.
    \item \textbf{TFV}: 
    The ground-truth is either ``entailed'' or ``refuted''. The reward is 1 if the predicted label matches the ground-truth, and 0 otherwise.
    \item \textbf{FF-TQA}: The ground-truth is a sentence or paragraph. We use the average of normalized BLEU and ROUGE-L scores to reflect semantic overlap between model output and reference answer.
\end{itemize}
This combination of rule-based and metric-based evaluation ensures reward interpretability and robustness, mitigating instability and reward hacking.

In addition to \emph{accuracy}, we encourage models—especially base models without instruction tuning—to follow a strict \emph{response format} specified by a system prompt. We introduce a \emph{cumulative format reward} incentivizing outputs that match the required template: <think>\ldots</think> <answer>\ldots</answer>, with the <answer> block containing a JSON snippet of the form \{"answer": ...\}. The format reward is computed by a deterministic regular expression-based checker, which assigns partial credit as outputs progressively satisfy structural requirements (\eg tag presence, JSON structure, valid answer types), and awards full credit for strictly conformant outputs. This dense and interpretable reward guides base models to generate well-structured, verifiable responses.

\section{Experiment}
In this section, we describe the experimental setup and address three central research questions, presenting the findings associated with each.

\subsection{\ours Model Training Details}
All models are trained using the verl framework. We initialize Table-R1 with Qwen2.5-7B-Instruct and Llama-3.1-8B-Instruct models. For Table-R1-SFT, training is conducted with a batch size of 256, a maximum sequence length of 20,480, and ulysses\_sequence\_parallel\_size set to 4. The learning rate is set to $1\times10^{-5}$, and training is performed for 3 epochs. For Table-R1-Zero, we use a batch size of 256 and 16 rollouts per prompt under the GRPO algorithm. It is trained for 2 epochs. The learning rate is fixed at $1\times10^{-6}$, sampling temperature is 1.0, with a maximum prompt length of 4096 tokens and a maximum response length of 1024 tokens. The GRPO clipping parameters are set to $\varepsilon_{\text{low}}=0.2$ and $\varepsilon_{\text{high}}=0.28$. For validation during training, inference is performed with temperature 0.6 and top-p 1.0. All the experiments are conducted on 4 NVIDIA A100 80GB GPUs.

\begin{table*}[!t]
\label{tab:main-results} 
\centering

\setlength{\tabcolsep}{2pt}   

\resizebox{\textwidth}{!}{
\begin{tabular}{l*{13}{c}}
\toprule

\multirow{3}{*}{\textbf{Model}} & \multicolumn{4}{c}{\textbf{In-domain Performance}} & \multicolumn{9}{c}{\textbf{Out-of-domain Performance}} \\
\cmidrule(lr){2-5} \cmidrule(lr){6-14}

& \multicolumn{1}{c}{\textbf{FF-TQA}} & \multicolumn{1}{c}{\textbf{TFV}} & \multicolumn{2}{c}{\textbf{TQA}} & \multicolumn{3}{c}{\textbf{FF-TQA}} & \multicolumn{3}{c}{\textbf{TFV}} & \multicolumn{3}{c}{\textbf{TQA}} \\
\cmidrule(lr){2-2} \cmidrule(lr){3-3} \cmidrule(lr){4-5} \cmidrule(lr){6-8} \cmidrule(lr){9-11} \cmidrule(lr){12-14}

& \multicolumn{1}{c}{FeTaQA} & \multicolumn{1}{c}{TabFact} & \multicolumn{1}{c}{WTQ} & \multicolumn{1}{c}{HiTab} & \multicolumn{1}{c}{ToTTo} & \multicolumn{1}{c}{QTSum} & \multicolumn{1}{c}{R.W.} & \multicolumn{1}{c}{InfoTabs} & \multicolumn{1}{c}{PHT} & \multicolumn{1}{c}{Feverous} & \multicolumn{1}{c}{TMCQ} & \multicolumn{1}{c}{TMWP} & \multicolumn{1}{c}{FinQA} \\
\midrule

\multicolumn{14}{c}{\textit{\textbf{Proprietary Models}}} \\ 
\addlinespace[0.5em]
GPT-4.1        & 25.1 & 86.5 & 68.0 & 84.7 & 20.4 & 45.7 & 21.0 & 90.5 & 88.2 & 87.7 & 92.0 & 77.0 & 74.0 \\
GPT-4.1 mini   & 27.2 & 84.9 & 69.5 & 80.7 & 18.8 & 46.4 & 20.0 & 88.9 & 86.8 & 86.1 & 92.9 & 86.2 & 71.4 \\
\midrule

\multicolumn{14}{c}{\textit{\textbf{Open-source Models}}} \\
\addlinespace[0.5em]

\multicolumn{14}{l}{\textit{LLMs}} \\
\quad Qwen2.5-7B   & 21.0 & 72.2 & 54.8 & 61.8 & 16.0 & 39.5 & 19.3 & 78.6 & 70.7 & 74.6 & 87.4 & 85.0 & 66.4 \\
\quad Qwen2.5-32B  & 21.9 & 90.3 & 77.3 & 79.4 & 17.8 & 41.4 & \underline{20.0} & 90.5 & 86.7 & 79.2 & 92.1 & 95.8 & \underline{77.3} \\
\quad DeepSeek-V3  & 24.7 & \textbf{92.4} & 69.9 & \underline{82.2} & 19.0 & \textbf{46.2} & \textbf{20.9} & \textbf{91.9} & 86.2 & \textbf{85.8} & 87.6 & 93.4 & \textbf{78.6} \\

\addlinespace[0.2em]
\cdashlinelr{1-14}
\addlinespace[0.5em]

\multicolumn{14}{l}{\textit{Reasoning LLMs}} \\
\addlinespace[0.2em]
\quad DeepSeek-R1-Distill-7B     & 19.1 & 79.6 & 57.8 & 46.2 & 10.7 & 37.2 & 18.0 & 85.7 & 87.1 & 77.5 & 80.9 & 94.0 & 66.8 \\
\quad QwQ-32B            & 23.8 & \underline{91.5} & \textbf{85.4} & 81.6 & 19.4 & 41.9 & 19.6 & \underline{91.0} & 87.8 & 80.1 & 90.7 & \textbf{99.4} & 76.2 \\
\quad DeepSeek-R1        & 26.2 & 90.8 & 79.6 & \textbf{82.4} & 18.5 & \underline{43.8} & 19.9 & 90.4 & 87.5 & 76.0 & \textbf{93.3} & \underline{99.0} & 75.8 \\

\addlinespace[0.2em]
\cdashlinelr{1-14}
\addlinespace[0.5em]

\multicolumn{14}{l}{\textit{Table-Specific LLMs}} \\
\addlinespace[0.2em]
\quad $^{\dagger}$TableBenchLLM (7B)     &  3.1 & 27.1 &  38.8 & -- & 6.2 & -- & -- & -- & -- & 42.3 & -- & -- & -- \\
\quad $^{\dagger}$TableLLM-13B    & 10.8 & 69.0  &  66.3 &  6.3 & 5.4 & -- & -- & -- & -- & 21.5 & -- & -- & -- \\
\quad $^{\dagger}$TableLlama (7B)     & \textbf{39.0} & 82.6 & 35.0 & 64.7 & 20.8 & -- & -- & -- & -- & 73.8 & -- & -- & -- \\ 
\quad TableGPT2-7B & 29.0 & 77.8 & 61.4 & 70.3 & 14.1 & 39.0 & 19.0 & 85.4 & \underline{89.1} & 78.0 & 77.2 & 79.7 & 66.4 \\ 
\quad $^{\dagger*}$TableGPT2-72B     & 32.3 & 85.4 & 71.5 & 75.6 & \textbf{22.7} & -- & -- & -- & -- & 76.8 & -- & -- & -- \\
\midrule

\multicolumn{14}{c}{\textit{\textbf{\ours (Ours)}}} \\
\addlinespace[0.5em]

\multicolumn{14}{l}{\textit{Llama-3.1 Series}} \\
\addlinespace[0.2em]

\quad Llama-3.1-8B-Instruct       & 21.7 & 74.1 & 52.3 & 58.2 & 16.5 & 31.6 & 18.1 & 84.1 & 82.5 & 78.3 & 49.5 & 72.0 & 57.1 \\
\quad Table-R1-SFT                 & 26.0 & 91.1 & \underline{83.8} & 81.8 & 13.7 & 36.6 & 16.6 & 89.8 & 85.8 & 79.4 & 90.8 & 89.0 & 64.3 \\
\quad Table-R1-Zero    & \underline{32.7} & 87.6 & 81.2 & 81.4 & \underline{22.3} & 30.2 & 17.7 & 87.9 & \textbf{91.6} & \underline{80.2}& 68.6 & 84.6 & 62.3 \\ 

\addlinespace[0.2em]
\cdashlinelr{1-14}
\addlinespace[0.5em]
\multicolumn{14}{l}{\textit{Qwen2.5 Series}} \\
\addlinespace[0.2em]

\quad Qwen2.5-7B-Instruct         & 21.0 & 72.2 & 54.8 & 61.8 & 16.0 & 39.5 & 19.3 & 78.6 & 70.7 & 74.6 & 87.4 & 85.0 & 66.4 \\
\quad Table-R1-SFT                 & 25.3 & 89.9 & 81.9 & 78.3 & 14.1 & 38.8 & 18.8 & 88.8 & 84.6 & 76.0 & 90.9 & 96.6 & 71.7 \\
\quad Table-R1-Zero                  & 30.6 & 87.6 & 79.8 & 78.1 & 19.8 & 43.1 & \underline{20.0} & 83.7 & 88.0 & 76.2 & \underline{93.0} & 96.4 & 70.8 \\
\bottomrule
\end{tabular}
}
\caption{Results on 13 table reasoning benchmarks spanning TQA, TFV, and FF-TQA tasks. For TQA, EM accuracy is reported (with ambiguous cases re-evaluated by GPT-4.1 mini); for TFV, classification accuracy; for FF-TQA, BLEU and ROUGE-L. \textbf{Bold} and \underline{underlined} scores indicate the top-2 performances among \emph{open-source} models.
$^{\dagger}$: Due to the context length limitations of most previous table-specific LLMs, it is challenging to conduct a fully fair comparison. Therefore, for these models, we directly use the results as reported in their respective papers, which may be based on sampled or filtered datasets. $^*$: Model weight has not been released.} 
\label{tab:main-results}
\end{table*}

\subsection{Experiment Setup}\label{sec:eval}
\paragraph{Evaluation Benchmarks.} 
To address the lack of a unified evaluation framework for table reasoning, we introduce a new benchmark suite encompassing a wide range of datasets.
For \emph{in-domain evaluation}, we use test sets from the same distributions as the training data: WTQ~\cite{pasupat-liang-2015-compositional} and HiTab~\cite{cheng-etal-2022-hitab} for TQA; TabFact~\cite{Chen2020TabFact} for TFV; and FeTaQA~\cite{nan-etal-2022-fetaqa} for FF-TQA. 
To assess generalization of \ours, we further conduct \emph{out-of-domain evaluation} on datasets that are not seen during training: TabMCQ~\cite{jauhar2016tabmcq}, TabMWP~\cite{lu2023dynamic}, and FinQA~\cite{chen-etal-2021-finqa} for TQA; InfoTabs~\cite{gupta-etal-2020-infotabs}, PubHealthTab~\cite{akhtar-etal-2022-pubhealthtab}, and Feverous~\cite{aly2021feverous} for TFV; ToTTo~\cite{parikh-etal-2020-totto}, QTSumm~\cite{zhao-etal-2023-qtsumm}, and RotoWire~\cite{wiseman-etal-2017-challenges} for FF-TQA. 
We provide detailed descriptions of the \textbf{evaluated datasets} and \textbf{evaluated  baseline systems} in Appendix~\ref{sec:appendix-dataset} and Appendix~\ref{app:exp-setup}.

\paragraph{Automated Evaluation System.}
For each evaluated dataset, we use its test set for evaluation.
For the TQA task, we report Exact Match (EM) accuracy. For TFV, we use classification accuracy. For FF-TQA, we measure generation quality with BLEU and ROUGE-L scores. Considering that EM accuracy in short-answer TQA may underestimate model performance due to formatting variations or semantically equivalent but non-exact matches, we further re-evaluate responses initially marked incorrect by EM using the GPT-4.1 mini model, with the prompt shown in Appendix~\ref{sec:appendix-prompt-llm-as-a-judge}.

\begin{tcolorbox}[colback=gray!10, colframe=gray!50!white, width=\linewidth, boxrule=0.5mm, arc=1mm, outer arc=1mm, left=2mm, right=2mm]
\textcolor{orange!50!yellow}{\faLightbulb} \xspace \textbf{RQ1:} How effective is inference-time scaling for table reasoning tasks?
\end{tcolorbox}

\noindent To address RQ1, we conduct a comprehensive evaluation and summarize the key findings from the results presented in \autoref{tab:main-results} as follows:

\subsection{Main Findings}
\paragraph{General-Purpose LLMs vs. Reasoning LLMs.}
We evaluate three pairs of general-purpose LLMs—with and without inference-time scaling capabilities—on table reasoning tasks: Qwen2.5-7B vs. DeepSeek-R1-Distill-7B, Qwen2.5-32B vs. QwQ-32B, and DeepSeek-V3 vs. DeepSeek-R1. While reasoning LLMs tend to outperform their counterparts on the TQA benchmark, their results on TFV and FF-TQA are mixed and not consistently superior. This indicates that inference-time scaling alone, without table-specific training, does not provide a clear advantage for table reasoning tasks. These findings underscore the importance of specialized adaptation strategies, such as Table-R1, for effective performance in this domain.

\paragraph{\ours In-Domain Performance.}
Both \sftmodel and \rlmodel achieve substantial improvements across all tasks. Specifically, Table-R1-Zero-8B obtains a BLEU score of 32.7 on FeTaQA for FF-TQA, significantly surpassing the best among other models of 26.2 from Deepseek-R1; Table-R1-SFT-8B reaches 91.1 accuracy on TabFact for TFV, closely matching the leading 91.9; and for TQA, our models achieve 83.8 and 81.8 on WTQ and HiTab, respectively, which are comparable to the best scores among other models of 85.4 and 84.7. These results consistently demonstrate that Table-R1 models deliver robust gains and competitive performance across diverse table reasoning scenarios, validating the effectiveness and versatility of both SFT and RLVR training strategies.

\paragraph{\ours Out-of-Domain Performance.}
Table-R1 models exhibit strong generalization capabilities on out-of-domain datasets. Across most out-of-domain benchmarks, our models consistently outperform their respective initial baselines, demonstrating the effectiveness of both SFT and R1-Zero training methods for table reasoning. Notably, Table-R1-Zero-7B achieves the best overall generalization among all variants. In contrast, we observe that supervised fine-tuning (SFT) leads to weaker generalization compared to the R1-Zero method, and models initialized from Llama tend to generalize less effectively than those based on Qwen. These results highlight the advantage of our approach in improving table reasoning robustness beyond the training distribution.

\begin{figure}[!t]
  \includegraphics[width=\columnwidth]{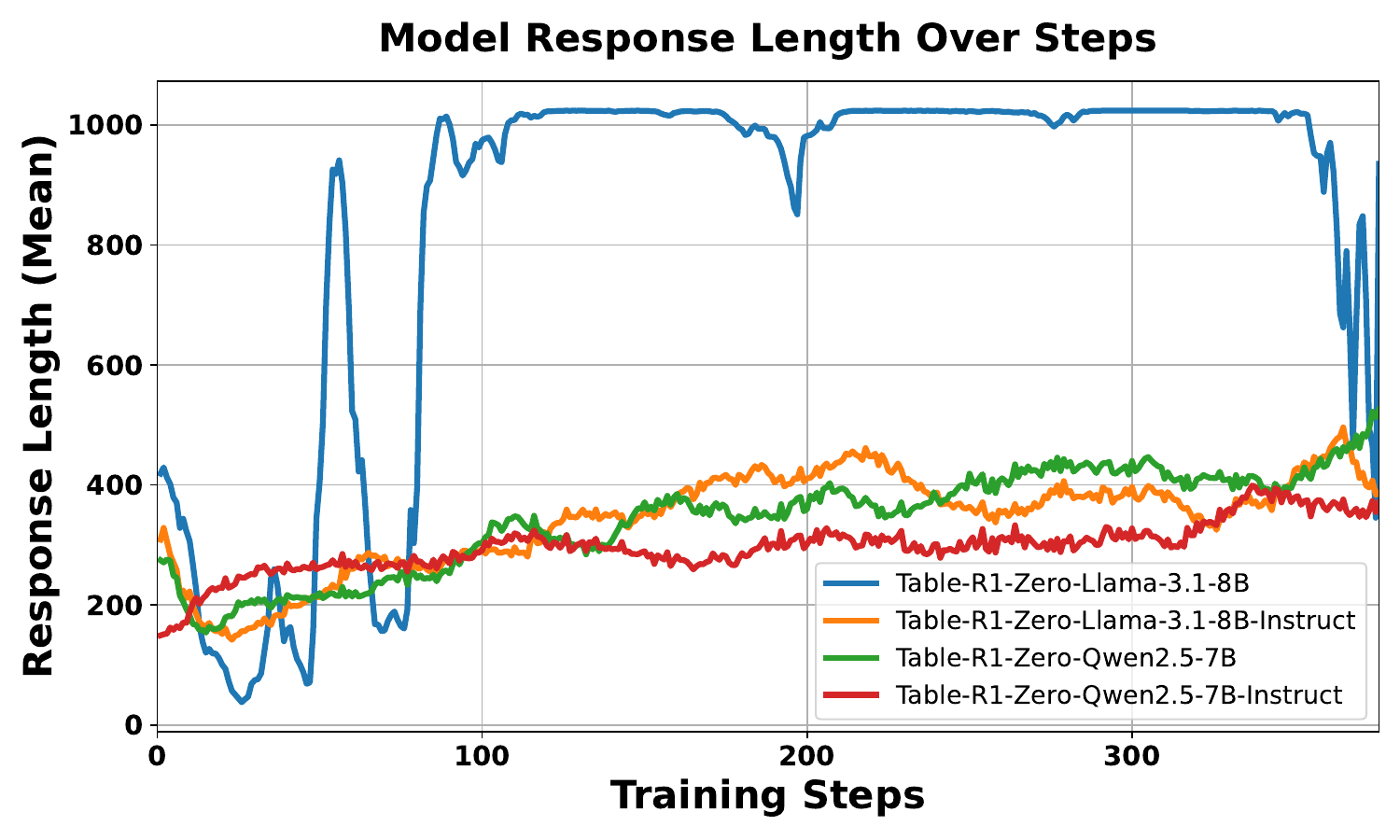}
  \caption{Response length during Table-R1 training across different models.}
  \label{fig:training-dynamics-response}
\end{figure}

\begin{figure*}[!t]
  \includegraphics[width=0.48\linewidth]{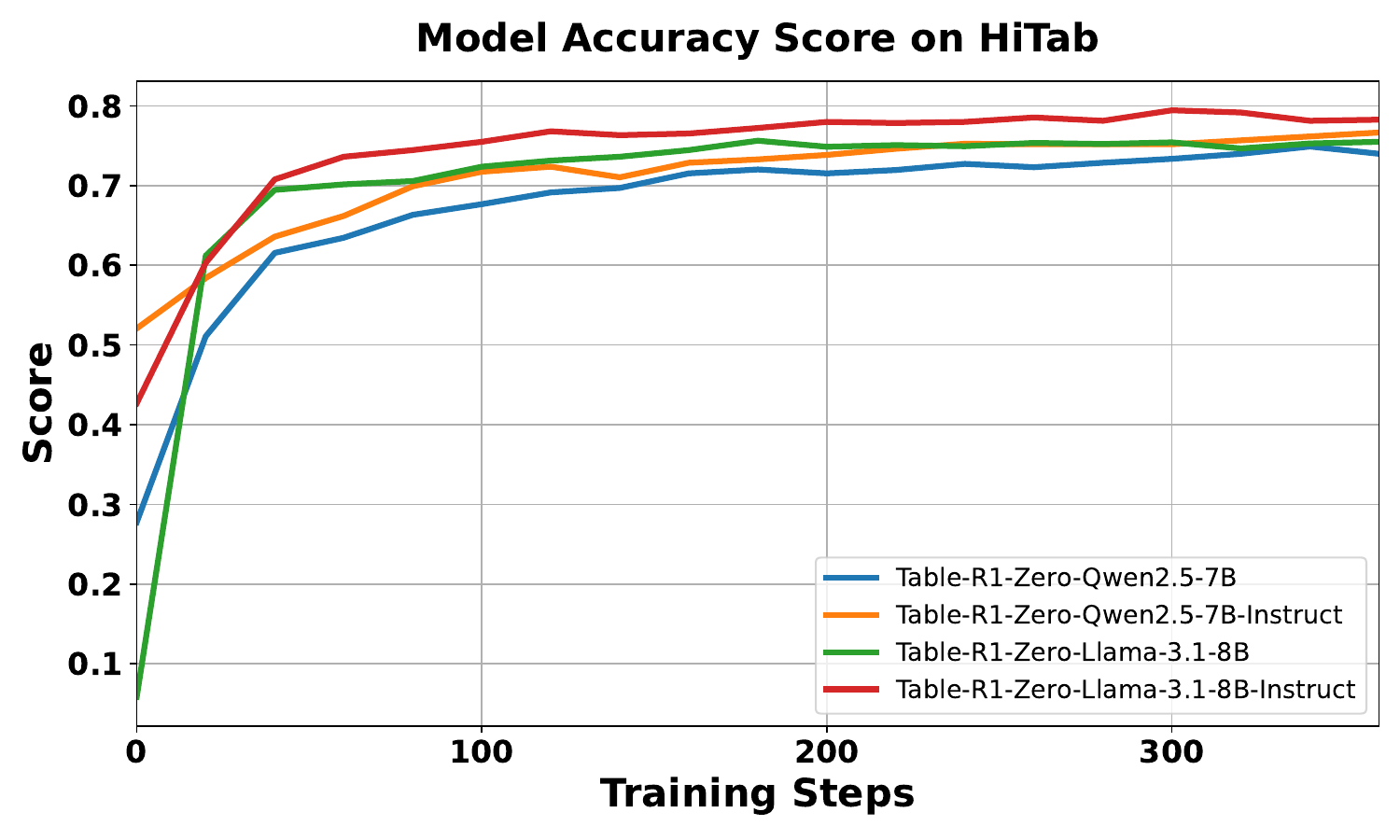} \hfill
  \includegraphics[width=0.48\linewidth]{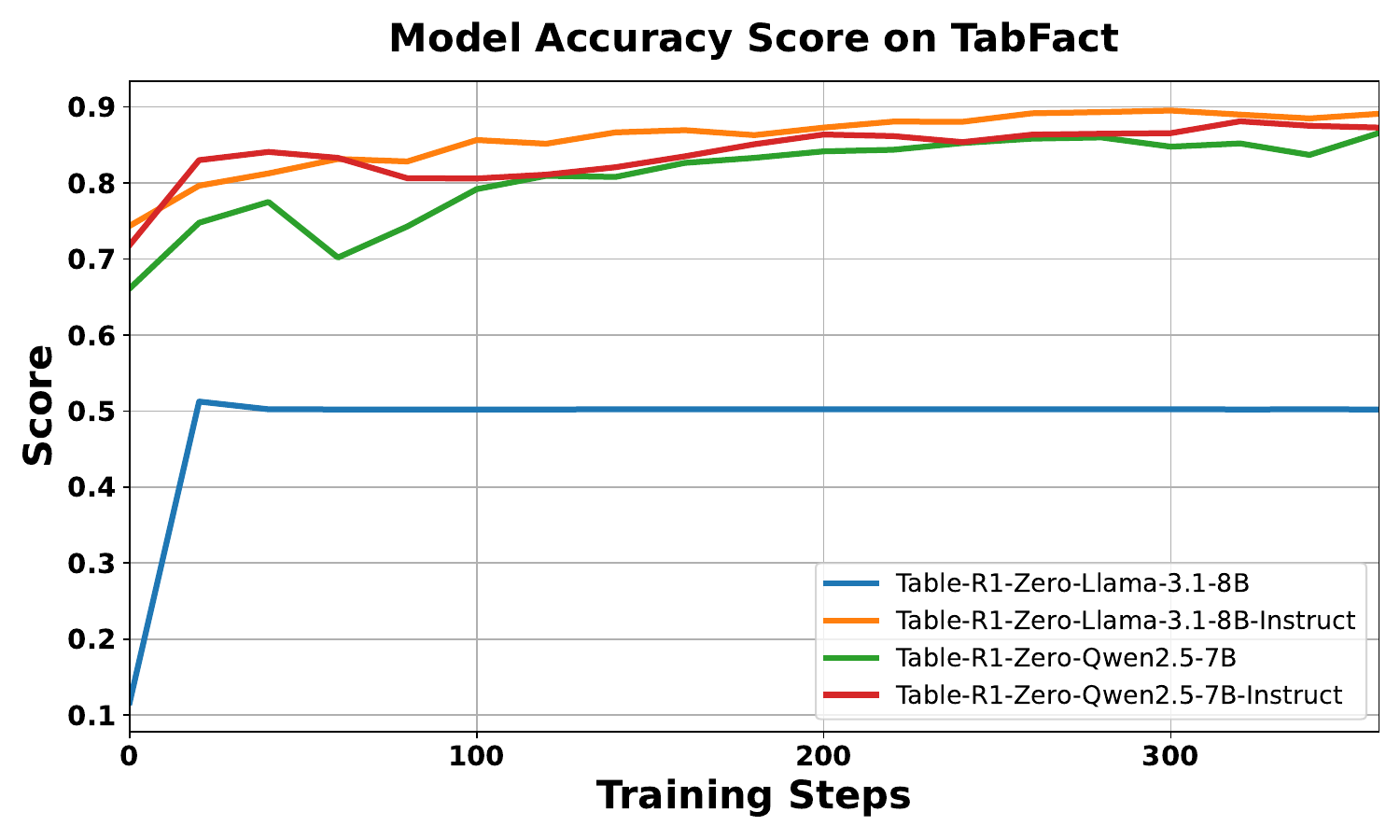} \hfill
  \includegraphics[width=0.48\linewidth]{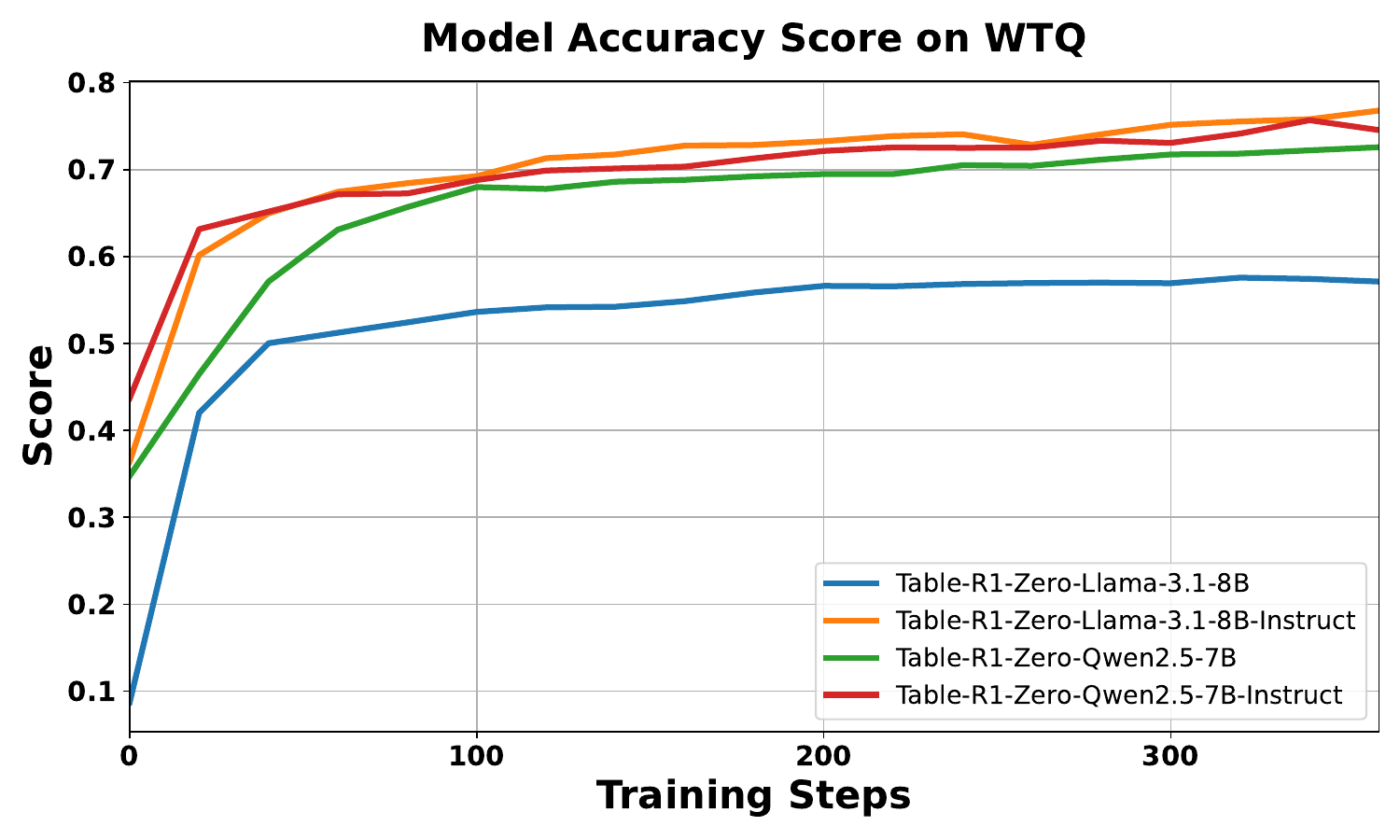} \hfill
  \includegraphics[width=0.48\linewidth]{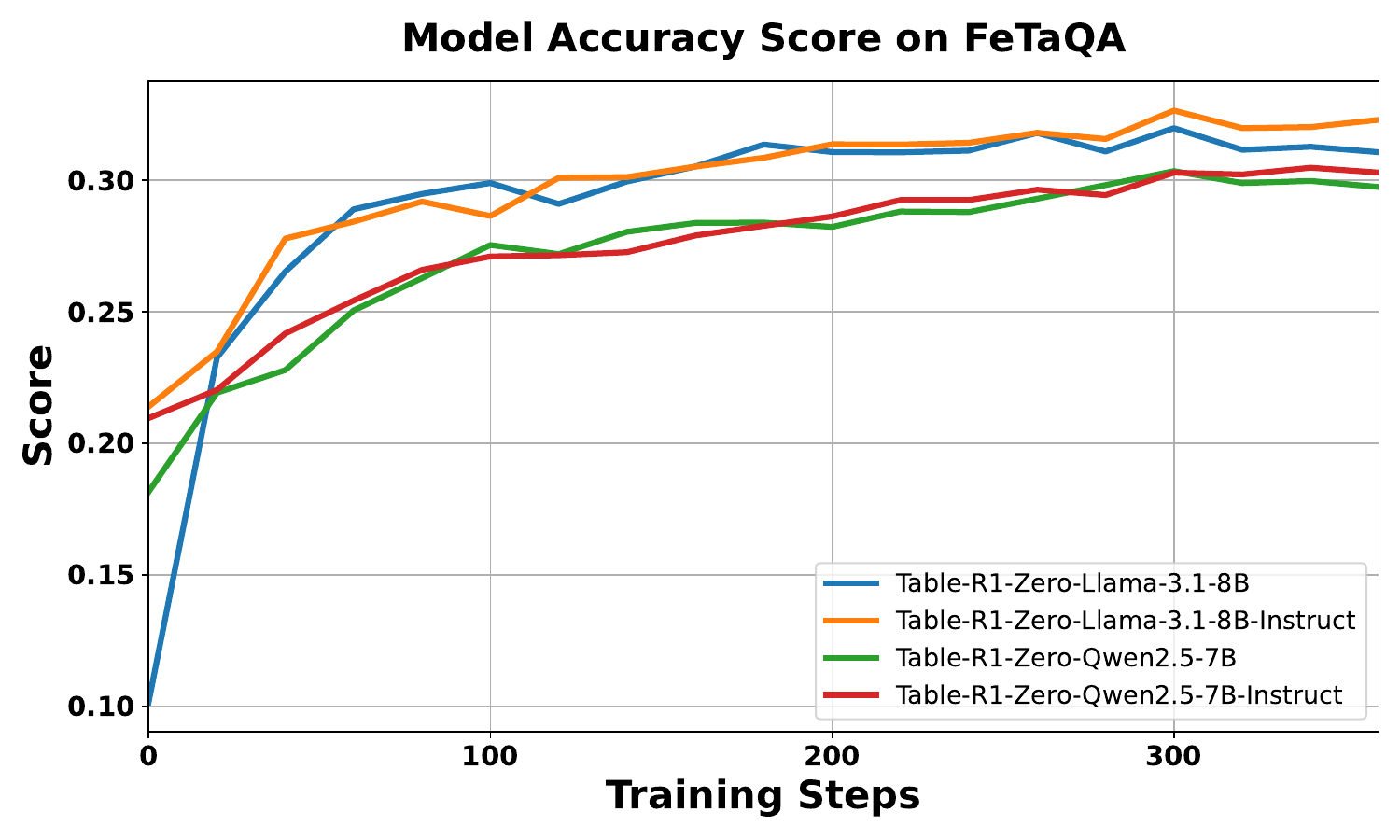}
\caption{Accuracy and BLEU score dynamics across four table reasoning datasets during RLVR training. Results are shown for all four Table-R1-Zero models, which are trained from Qwen2.5 7B or Llama-3.1 8B as initialization.}
  \label{fig:training-dynamics-accuracy}
\end{figure*}

\begin{tcolorbox}[colback=gray!10, colframe=gray!50!white, width=\linewidth, boxrule=0.5mm, arc=1mm, outer arc=1mm, left=2mm, right=2mm]
\textcolor{orange!50!yellow}{\faLightbulb} \xspace \textbf{RQ2:} What contributes to the success of RLVR methods in table reasoning tasks?
\end{tcolorbox}

\noindent To address this research question, we present a comprehensive analysis of \rlmodel in the following three subsections: the training dynamics, a qualitative assessment of model responses, and an exploration of the reasoning capacity boundaries.

\subsection{Analysis of Training Dynamics}\label{sec:dynamics}
We conduct a detailed analysis of the training dynamics exhibited by our Table-R1 models across various model backbones, including Qwen2.5 7B and Llama-3.1 8B, under both \textit{base} and \textit{instruct} configurations.

\autoref{fig:training-dynamics-response} presents the evolution of \textit{response length} throughout reinforcement learning. Notably, the base models consistently start with longer responses compared to their instruct counterparts. During the initial stage of RL training, we observe a sharp drop in response length, corresponding to a phase of format acquisition—where the model learns to produce outputs adhering to the expected answer format. Subsequently, response length gradually increases, with base models exhibiting a more pronounced growth trajectory than instruct models. Among all four model variants, Table-R1-Zero-Qwen2.5-7B-Instruct demonstrates the greatest stability, showing a smooth and moderate increase in response length. In contrast, Table-R1-Zero-Llama-3.1-8B displays considerable instability, ultimately failing to acquire the desired response format. We attribute this to the weaker instruction-following capabilities of the Llama-3.1-8B base model.

\autoref{fig:training-dynamics-accuracy} illustrates the progression of \textit{model accuracy} over the course of RL training. We find that instruct models consistently achieve higher accuracy than their base versions throughout training. While base models start from a lower performance baseline, they undergo a phase of format adaptation, after which their accuracy trends converge with those of instruct models. Interestingly, although Table-R1-Zero-Llama-3.1-8B exhibits unstable changes in response length, it still demonstrates performance gains on most tasks, with the exception of TabFact—a binary classification task.

Overall, these observations reveal that instruction tuning provides significant benefits in both stability and performance during RLVR training for table reasoning tasks. The results further suggest that initial format alignment and instruction-following capability play critical roles in the successful adaptation of large language models to structured reasoning scenarios.

\subsection{Reasoning Capacity Boundary}
\begin{figure*}[!t]
  \includegraphics[width=\linewidth]{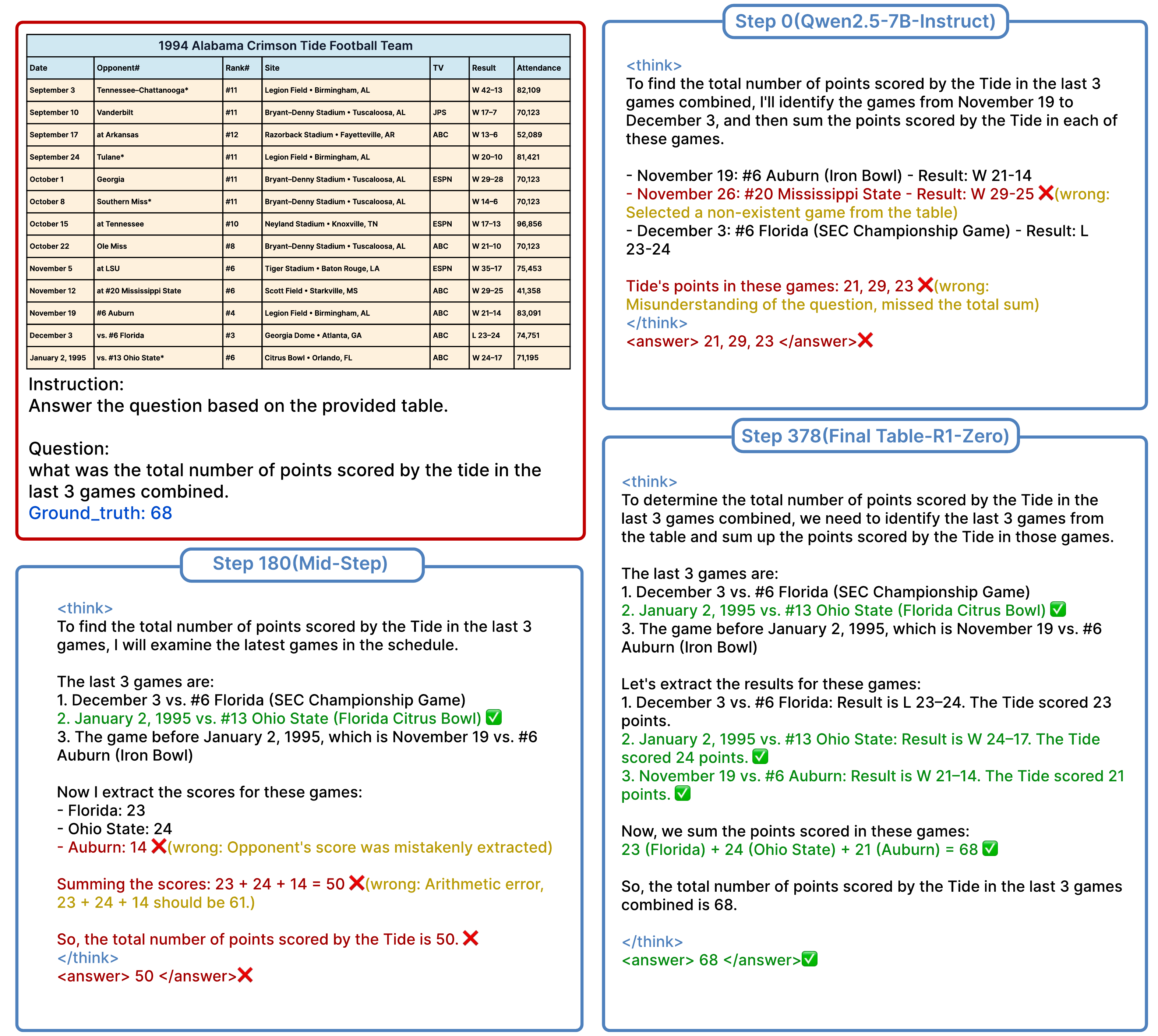}
  \caption{Illustration of the model's reasoning progression across training steps. The example demonstrates how reasoning quality evolves from superficial processing (Step 0), to partial column-aware reasoning (Step 180), and finally to accurate multi-step inference with semantic and arithmetic understanding (Step 378).}
  \label{fig:qualitative-analysis}
\end{figure*}

Understanding the upper limits of a model's reasoning ability is crucial for evaluating the true impact of RLVR on tabular reasoning tasks. Inspired by prior studies on the boundaries of RLVR method~\cite{yue2025does}, we employ the \textit{pass@k} metric to quantify the model's capacity. Specifically, pass@k measures the probability that at least one out of $k$ generated responses is correct, given a fixed input prompt. This metric is particularly well-suited for our setting, as it captures not only the model's accuracy, but also its ability to produce diverse and plausible reasoning trajectories within a limited number of attempts.

We systematically evaluate pass@k for $k$ up to 32, both before and after RLVR training, on the HiTab and WTQ datasets. As illustrated in \autoref{fig:passk}, RLVR training leads to a notable increase in pass@k values throughout the evaluated range. This improvement is consistent across different datasets and model architectures, demonstrating that RLVR enhances not only the likelihood of obtaining a correct answer on first attempt (pass@1), but also the breadth of valid reasoning paths the model can explore in a small sampling budget.

\subsection{Qualitative Analysis}
To gain deeper insights into how RLVR shapes model behavior, we conduct a qualitative analysis of model responses throughout the training process. By examining the same set of representative prompts across different RL training steps, we observe that the model not only internalizes general R1-style reasoning characteristic, but also acquires table-specific reasoning abilities critical for tabular tasks, as illustrated in \autoref{fig:qualitative-analysis} and Appendix~\ref{app:example}.

On the \textbf{reasoning} axis, we observe clear progress toward sophisticated, multi-step reasoning: after RL training, the \rlmodel model decomposes complex queries into sequential sub-tasks, explicitly outlining intermediate steps and sometimes ``rethinking'' earlier conclusions to check or correct errors. Such reflective patterns, rare at initialization, become prevalent with training, suggesting the verifiable reward encourages explicit, auditable reasoning.

On the \textbf{table-specific} axis, we observe notable improvements in three key areas. First, the model develops column-aware reasoning: it accurately identifies and references relevant table columns, often justifying its answer with explicit column mentions or by highlighting how information from multiple columns is synthesized. Second, the model demonstrates enhanced semantic understanding of natural language questions, especially in TQA settings. It is able to correctly interpret nuanced question intents (\eg comparative, aggregative, or conditional queries) and map them to the corresponding table structures. Third, we note a marked increase in arithmetic and temporal reasoning capabilities. The model becomes more adept at performing arithmetic calculations over table entries and reasoning over temporal sequences, both of which are crucial for table reasoning tasks.

\subsection{Ablation Studies on RLVR Training}
\begin{tcolorbox}[colback=gray!10, colframe=gray!50!white, width=\linewidth, boxrule=0.5mm, arc=1mm, outer arc=1mm, left=2mm, right=2mm]
\textcolor{orange!50!yellow}{\faLightbulb} \xspace \textbf{RQ3:} How do various factors influence the effectiveness of RLVR training in table reasoning tasks?
\end{tcolorbox}

\noindent To better understand the contributions of both SFT and RLVR methods and to assess the robustness of Table-R1 across different configurations, we conduct extensive ablation studies from three perspectives: model-level, task-level, and explicit reasoning format. We detail our findings as follows.

\paragraph{Model-level Ablation.}
We first analyze the effect of model initialization by comparing base and instruct variants. Across all settings, instruct models consistently outperform their base counterparts. We attribute this to the enhanced instruction-following ability inherent to instruct models, which enables faster adaptation to the explicit reasoning formats required by table tasks and leads to more stable training dynamics. 

We next compare representative model architectures—Qwen2.5 and Llama-3.1. Under identical training regimes, Llama-3.1-8B-Instruct achieves superior in-domain performance on table reasoning benchmarks compared to Qwen2.5-7B-Instruct, indicating a stronger capacity for learning table-specific reasoning. However, Qwen2.5 demonstrates better out-of-domain generalization, suggesting that model architecture and pretraining data may influence the balance between in-domain accuracy and cross-domain robustness.

Furthermore, we evaluate the effect of distillation data by comparing our SFT models—fine-tuned on table-specific DeepSeek-R1 distillation data—against the official DeepSeek-R1 distilled models. Our models not only surpass the official versions, but also outperform larger-scale distilled models. This underscores the effectiveness of domain-specific fine-tuning and the importance of high-quality, task-aligned training data.

\paragraph{Task-level Ablation.}
To investigate cross-task generalization, we train Table-R1 exclusively on the TQA dataset and evaluate its performance on TFV and FF-TQA tasks. Interestingly, models trained solely on TQA exhibit notable performance gains on TFV, indicating that the reasoning capabilities required for TFV are closely aligned with those developed for short-answer TQA. In contrast, no significant improvement is observed on FF-TQA, likely due to the distinct reasoning and answer generation skills required for free-form responses, which are not adequately covered by TQA training. These results highlight the varying degrees of transferability among table reasoning tasks and emphasize the need for targeted training to achieve robust generalization.

\paragraph{Format Ablation.}
We assess the role of explicit reasoning format supervision by removing the format reward during training. This ablation reduces training stability and slightly lowers in-domain performance. More notably, generalization suffers: while TFV scores may improve, performance on short-answer TQA and FF-TQA declines, sometimes even below baseline, indicating that format supervision is crucial for transferable reasoning.

\section{Conclusion}
This work presents the first comprehensive study on applying inference-time scaling to table reasoning tasks. 
Through extensive evaluation across 13 diverse benchmarks, we demonstrate that inference-time scaling enables substantial improvements in reasoning quality, with RLVR methods yielding stronger generalization to out-of-domain tasks. 
Ablation studies confirm the benefits of instruction tuning, model architecture choice, and task composition in enhancing training effectiveness. Qualitative analysis reveals that RLVR fosters the emergence of structured, multi-step reasoning and table-specific capabilities. We hope this work paves the way for future research in structured reasoning.

\section*{Limitations}
Several limitations remain that warrant future investigation:
The SFT data was generated exclusively using DeepSeek-R1. Additionally, the data verification and filtering processes may have inadvertently removed difficult or high-quality training examples. Future research could explore incorporating outputs from other reasoning LLMs, such as QwQ-32B, to enhance distillation performance and data diversity.
Furthermore, in Section~\ref{sec:dynamics}, we observe that models initialized from the LLaMA-3.1-8B backbone exhibit unstable training dynamics during RLVR fine-tuning, including inconsistent acquisition of the desired output format and significant fluctuations in response length. While instruct-tuned variants mitigate some of these issues, the underlying causes of instability—such as sensitivity to initialization, reward sparsity, or optimization hyperparameters—remain underexplored.
Future work could investigate strategies to improve the robustness and generalizability of RLVR for structured reasoning tasks.

\section*{Acknowledgments}
We would like to thank Zijie Zhou for making the TQA-Distill-R1 dataset\footnote{\url{https://huggingface.co/datasets/jared-zhou/TQA-Distill-R1}} publicly available, which we used in our early experiments.

\bibliography{anthology, custom}

\begin{thebibliography}{75}
\providecommand{\natexlab}[1]{#1}

\bibitem[{Akhtar et~al.(2022)Akhtar, Cocarascu, and Simperl}]{akhtar-etal-2022-pubhealthtab}
Mubashara Akhtar, Oana Cocarascu, and Elena Simperl. 2022.
\newblock \href {https://doi.org/10.18653/v1/2022.findings-naacl.1} {{P}ub{H}ealth{T}ab: {A} public health table-based dataset for evidence-based fact checking}.
\newblock In \emph{Findings of the Association for Computational Linguistics: NAACL 2022}, pages 1--16, Seattle, United States. Association for Computational Linguistics.

\bibitem[{Aly et~al.(2021)Aly, Guo, Schlichtkrull, Thorne, Vlachos, Christodoulopoulos, Cocarascu, and Mittal}]{aly2021feverous}
Rami Aly, Zhijiang Guo, Michael~Sejr Schlichtkrull, James Thorne, Andreas Vlachos, Christos Christodoulopoulos, Oana Cocarascu, and Arpit Mittal. 2021.
\newblock \href {https://openreview.net/forum?id=h-flVCIlstW} {{FEVEROUS}: Fact extraction and {VER}ification over unstructured and structured information}.
\newblock In \emph{Thirty-fifth Conference on Neural Information Processing Systems Datasets and Benchmarks Track (Round 1)}.

\bibitem[{Chen et~al.(2020)Chen, Wang, Chen, Zhang, Wang, Li, Zhou, and Wang}]{Chen2020TabFact}
Wenhu Chen, Hongmin Wang, Jianshu Chen, Yunkai Zhang, Hong Wang, Shiyang Li, Xiyou Zhou, and William~Yang Wang. 2020.
\newblock \href {https://openreview.net/forum?id=rkeJRhNYDH} {Tabfact: A large-scale dataset for table-based fact verification}.
\newblock In \emph{International Conference on Learning Representations}.

\bibitem[{Chen et~al.(2021)Chen, Chen, Smiley, Shah, Borova, Langdon, Moussa, Beane, Huang, Routledge, and Wang}]{chen-etal-2021-finqa}
Zhiyu Chen, Wenhu Chen, Charese Smiley, Sameena Shah, Iana Borova, Dylan Langdon, Reema Moussa, Matt Beane, Ting-Hao Huang, Bryan Routledge, and William~Yang Wang. 2021.
\newblock \href {https://doi.org/10.18653/v1/2021.emnlp-main.300} {{F}in{QA}: A dataset of numerical reasoning over financial data}.
\newblock In \emph{Proceedings of the 2021 Conference on Empirical Methods in Natural Language Processing}, pages 3697--3711, Online and Punta Cana, Dominican Republic. Association for Computational Linguistics.

\bibitem[{Cheng et~al.(2022)Cheng, Dong, Wang, Jia, Guo, Gao, Han, Lou, and Zhang}]{cheng-etal-2022-hitab}
Zhoujun Cheng, Haoyu Dong, Zhiruo Wang, Ran Jia, Jiaqi Guo, Yan Gao, Shi Han, Jian-Guang Lou, and Dongmei Zhang. 2022.
\newblock \href {https://doi.org/10.18653/v1/2022.acl-long.78} {{H}i{T}ab: A hierarchical table dataset for question answering and natural language generation}.
\newblock In \emph{Proceedings of the 60th Annual Meeting of the Association for Computational Linguistics (Volume 1: Long Papers)}, pages 1094--1110, Dublin, Ireland. Association for Computational Linguistics.

\bibitem[{DeepSeek-AI et~al.(2025)DeepSeek-AI, Guo, Yang, Zhang, Song, Zhang, Xu, Zhu, Ma, Wang, Bi, Zhang, Yu, Wu, Wu, Gou, Shao, Li, Gao, Liu, Xue, Wang, Wu, Feng, Lu, Zhao, Deng, Zhang, Ruan, Dai, Chen, Ji, Li, Lin, Dai, Luo, Hao, Chen, Li, Zhang, Bao, Xu, Wang, Ding, Xin, Gao, Qu, Li, Guo, Li, Wang, Chen, Yuan, Qiu, Li, Cai, Ni, Liang, Chen, Dong, Hu, Gao, Guan, Huang, Yu, Wang, Zhang, Zhao, Wang, Zhang, Xu, Xia, Zhang, Zhang, Tang, Li, Wang, Li, Tian, Huang, Zhang, Wang, Chen, Du, Ge, Zhang, Pan, Wang, Chen, Jin, Chen, Lu, Zhou, Chen, Ye, Wang, Yu, Zhou, Pan, Li, Zhou, Wu, Ye, Yun, Pei, Sun, Wang, Zeng, Zhao, Liu, Liang, Gao, Yu, Zhang, Xiao, An, Liu, Wang, Chen, Nie, Cheng, Liu, Xie, Liu, Yang, Li, Su, Lin, Li, Jin, Shen, Chen, Sun, Wang, Song, Zhou, Wang, Shan, Li, Wang, Wei, Zhang, Xu, Li, Zhao, Sun, Wang, Yu, Zhang, Shi, Xiong, He, Piao, Wang, Tan, Ma, Liu, Guo, Ou, Wang, Gong, Zou, He, Xiong, Luo, You, Liu, Zhou, Zhu, Xu, Huang, Li, Zheng, Zhu, Ma, Tang, Zha, Yan, Ren, Ren, Sha, Fu, Xu, Xie, Zhang,
  Hao, Ma, Yan, Wu, Gu, Zhu, Liu, Li, Xie, Song, Pan, Huang, Xu, Zhang, and Zhang}]{deepseekr1}
DeepSeek-AI, Daya Guo, Dejian Yang, Haowei Zhang, Junxiao Song, Ruoyu Zhang, Runxin Xu, Qihao Zhu, Shirong Ma, Peiyi Wang, Xiao Bi, Xiaokang Zhang, Xingkai Yu, Yu~Wu, Z.~F. Wu, Zhibin Gou, Zhihong Shao, Zhuoshu Li, Ziyi Gao, and 181 others. 2025.
\newblock \href {https://arxiv.org/abs/2501.12948} {Deepseek-r1: Incentivizing reasoning capability in llms via reinforcement learning}.
\newblock \emph{Preprint}, arXiv:2501.12948.

\bibitem[{Deng et~al.(2024)Deng, Sun, He, Sikka, Chen, Ma, Zhang, and Mihalcea}]{deng2024tables}
Naihao Deng, Zhenjie Sun, Ruiqi He, Aman Sikka, Yulong Chen, Lin Ma, Yue Zhang, and Rada Mihalcea. 2024.
\newblock Tables as texts or images: Evaluating the table reasoning ability of llms and mllms.
\newblock \emph{arXiv preprint arXiv:2402.12424}.

\bibitem[{Deng et~al.(2025)Deng, Zhang, Zhu, Chang, Zhang, Li, Hang, Kobayashi, Hu, and Ng}]{deng2025betterunderstandingtableinstruction}
Naihao Deng, Sheng Zhang, Henghui Zhu, Shuaichen Chang, Jiani Zhang, Alexander~Hanbo Li, Chung-Wei Hang, Hideo Kobayashi, Yiqun Hu, and Patrick Ng. 2025.
\newblock \href {https://arxiv.org/abs/2501.14717} {Towards better understanding table instruction tuning: Decoupling the effects from data versus models}.
\newblock \emph{Preprint}, arXiv:2501.14717.

\bibitem[{Face(2025)}]{openr1}
Hugging Face. 2025.
\newblock \href {https://github.com/huggingface/open-r1} {Open r1: A fully open reproduction of deepseek-r1}.

\bibitem[{Feng et~al.(2025{\natexlab{a}})Feng, Huang, Qu, Zhang, Qin, Zhong, Jiang, Chi, and Zhong}]{feng2025retool}
Jiazhan Feng, Shijue Huang, Xingwei Qu, Ge~Zhang, Yujia Qin, Baoquan Zhong, Chengquan Jiang, Jinxin Chi, and Wanjun Zhong. 2025{\natexlab{a}}.
\newblock Retool: Reinforcement learning for strategic tool use in llms.
\newblock \emph{arXiv preprint arXiv:2504.11536}.

\bibitem[{Feng et~al.(2023)Feng, Wan, Wen, McAleer, Wen, Zhang, and Wang}]{feng2023alphazero}
Xidong Feng, Ziyu Wan, Muning Wen, Stephen~Marcus McAleer, Ying Wen, Weinan Zhang, and Jun Wang. 2023.
\newblock Alphazero-like tree-search can guide large language model decoding and training.
\newblock \emph{arXiv preprint arXiv:2309.17179}.

\bibitem[{Feng et~al.(2025{\natexlab{b}})Feng, Ren, Su, Zheng, Tang, Wang, and Liu}]{feng2025mt}
Zhaopeng Feng, Jiahan Ren, Jiayuan Su, Jiamei Zheng, Zhihang Tang, Hongwei Wang, and Zuozhu Liu. 2025{\natexlab{b}}.
\newblock Mt-rewardtree: A comprehensive framework for advancing llm-based machine translation via reward modeling.
\newblock \emph{arXiv preprint arXiv:2503.12123}.

\bibitem[{Guan et~al.(2025)Guan, Zhang, Liu, Shang, Sun, Zhu, Yang, and Yang}]{guan2025rstar}
Xinyu Guan, Li~Lyna Zhang, Yifei Liu, Ning Shang, Youran Sun, Yi~Zhu, Fan Yang, and Mao Yang. 2025.
\newblock rstar-math: Small llms can master math reasoning with self-evolved deep thinking.
\newblock \emph{arXiv preprint arXiv:2501.04519}.

\bibitem[{Guo et~al.(2025)Guo, Yang, Zhang, Song, Zhang, Xu, Zhu, Ma, Wang, Bi et~al.}]{guo2025deepseek}
Daya Guo, Dejian Yang, Haowei Zhang, Junxiao Song, Ruoyu Zhang, Runxin Xu, Qihao Zhu, Shirong Ma, Peiyi Wang, Xiao Bi, and 1 others. 2025.
\newblock Deepseek-r1: Incentivizing reasoning capability in llms via reinforcement learning.
\newblock \emph{arXiv preprint arXiv:2501.12948}.

\bibitem[{Gupta et~al.(2020)Gupta, Mehta, Nokhiz, and Srikumar}]{gupta-etal-2020-infotabs}
Vivek Gupta, Maitrey Mehta, Pegah Nokhiz, and Vivek Srikumar. 2020.
\newblock \href {https://doi.org/10.18653/v1/2020.acl-main.210} {{INFOTABS}: Inference on tables as semi-structured data}.
\newblock In \emph{Proceedings of the 58th Annual Meeting of the Association for Computational Linguistics}, pages 2309--2324, Online. Association for Computational Linguistics.

\bibitem[{Handler et~al.(2024)Handler, Larsen, and Hackathorn}]{handler2024large}
Abram Handler, Kai~R Larsen, and Richard Hackathorn. 2024.
\newblock Large language models present new questions for decision support.
\newblock \emph{International Journal of Information Management}, 79:102811.

\bibitem[{Herzig et~al.(2020)Herzig, Nowak, M{\"u}ller, Piccinno, and Eisenschlos}]{herzig-etal-2020-tapas}
Jonathan Herzig, Pawel~Krzysztof Nowak, Thomas M{\"u}ller, Francesco Piccinno, and Julian Eisenschlos. 2020.
\newblock \href {https://doi.org/10.18653/v1/2020.acl-main.398} {{T}a{P}as: Weakly supervised table parsing via pre-training}.
\newblock In \emph{Proceedings of the 58th Annual Meeting of the Association for Computational Linguistics}, pages 4320--4333, Online. Association for Computational Linguistics.

\bibitem[{Hu et~al.(2025)Hu, Zhang, Han, Jiang, Zhang, and Shum}]{hu2025open}
Jingcheng Hu, Yinmin Zhang, Qi~Han, Daxin Jiang, Xiangyu Zhang, and Heung-Yeung Shum. 2025.
\newblock Open-reasoner-zero: An open source approach to scaling up reinforcement learning on the base model.
\newblock \emph{arXiv preprint arXiv:2503.24290}.

\bibitem[{Huang et~al.(2025{\natexlab{a}})Huang, Jia, Zhai, Cao, Ye, Zhao, Xu, Hu, and Lin}]{huang2025visionr1}
Wenxuan Huang, Bohan Jia, Zijie Zhai, Shaosheng Cao, Zheyu Ye, Fei Zhao, Zhe Xu, Yao Hu, and Shaohui Lin. 2025{\natexlab{a}}.
\newblock \href {https://arxiv.org/abs/2503.06749} {Vision-r1: Incentivizing reasoning capability in multimodal large language models}.
\newblock \emph{Preprint}, arXiv:2503.06749.

\bibitem[{Huang et~al.(2025{\natexlab{b}})Huang, Jia, Zhai, Cao, Ye, Zhao, Xu, Hu, and Lin}]{huang2025vision}
Wenxuan Huang, Bohan Jia, Zijie Zhai, Shaosheng Cao, Zheyu Ye, Fei Zhao, Zhe Xu, Yao Hu, and Shaohui Lin. 2025{\natexlab{b}}.
\newblock Vision-r1: Incentivizing reasoning capability in multimodal large language models.
\newblock \emph{arXiv preprint arXiv:2503.06749}.

\bibitem[{Jaech et~al.(2024)Jaech, Kalai, Lerer, Richardson, El-Kishky, Low, Helyar, Madry, Beutel, Carney et~al.}]{jaech2024openai}
Aaron Jaech, Adam Kalai, Adam Lerer, Adam Richardson, Ahmed El-Kishky, Aiden Low, Alec Helyar, Aleksander Madry, Alex Beutel, Alex Carney, and 1 others. 2024.
\newblock Openai o1 system card.
\newblock \emph{arXiv preprint arXiv:2412.16720}.

\bibitem[{Jauhar et~al.(2016)Jauhar, Turney, and Hovy}]{jauhar2016tabmcq}
Sujay~Kumar Jauhar, Peter Turney, and Eduard Hovy. 2016.
\newblock \href {https://arxiv.org/abs/1602.03960} {Tabmcq: A dataset of general knowledge tables and multiple-choice questions}.
\newblock \emph{Preprint}, arXiv:1602.03960.

\bibitem[{Jin et~al.(2025)Jin, Zeng, Yue, Yoon, Arik, Wang, Zamani, and Han}]{jin2025searchr1}
Bowen Jin, Hansi Zeng, Zhenrui Yue, Jinsung Yoon, Sercan Arik, Dong Wang, Hamed Zamani, and Jiawei Han. 2025.
\newblock \href {https://arxiv.org/abs/2503.09516} {Search-r1: Training llms to reason and leverage search engines with reinforcement learning}.
\newblock \emph{Preprint}, arXiv:2503.09516.

\bibitem[{Liang et~al.(2024)Liang, Zhang, Wu, Lepp, Ji, Zhao, Cao, Liu, He, Huang et~al.}]{liang2024mapping}
Weixin Liang, Yaohui Zhang, Zhengxuan Wu, Haley Lepp, Wenlong Ji, Xuandong Zhao, Hancheng Cao, Sheng Liu, Siyu He, Zhi Huang, and 1 others. 2024.
\newblock Mapping the increasing use of llms in scientific papers.
\newblock \emph{arXiv preprint arXiv:2404.01268}.

\bibitem[{Lightman et~al.(2023)Lightman, Kosaraju, Burda, Edwards, Baker, Lee, Leike, Schulman, Sutskever, and Cobbe}]{lightman2023let}
Hunter Lightman, Vineet Kosaraju, Yuri Burda, Harrison Edwards, Bowen Baker, Teddy Lee, Jan Leike, John Schulman, Ilya Sutskever, and Karl Cobbe. 2023.
\newblock Let's verify step by step.
\newblock In \emph{The Twelfth International Conference on Learning Representations}.

\bibitem[{Liu et~al.(2022)Liu, Chen, Guo, Ziyadi, Lin, Chen, and Lou}]{liu2022tapex}
Qian Liu, Bei Chen, Jiaqi Guo, Morteza Ziyadi, Zeqi Lin, Weizhu Chen, and Jian-Guang Lou. 2022.
\newblock \href {https://openreview.net/forum?id=O50443AsCP} {{TAPEX}: Table pre-training via learning a neural {SQL} executor}.
\newblock In \emph{International Conference on Learning Representations}.

\bibitem[{Liu et~al.(2025)Liu, Chen, Li, Qi, Pang, Du, Lee, and Lin}]{liu2025understanding}
Zichen Liu, Changyu Chen, Wenjun Li, Penghui Qi, Tianyu Pang, Chao Du, Wee~Sun Lee, and Min Lin. 2025.
\newblock Understanding r1-zero-like training: A critical perspective.
\newblock \emph{arXiv preprint arXiv:2503.20783}.

\bibitem[{Lu et~al.(2023)Lu, Qiu, Chang, Wu, Zhu, Rajpurohit, Clark, and Kalyan}]{lu2023dynamic}
Pan Lu, Liang Qiu, Kai-Wei Chang, Ying~Nian Wu, Song-Chun Zhu, Tanmay Rajpurohit, Peter Clark, and Ashwin Kalyan. 2023.
\newblock \href {https://openreview.net/forum?id=DHyHRBwJUTN} {Dynamic prompt learning via policy gradient for semi-structured mathematical reasoning}.
\newblock In \emph{The Eleventh International Conference on Learning Representations}.

\bibitem[{Muennighoff et~al.(2025)Muennighoff, Yang, Shi, Li, Fei-Fei, Hajishirzi, Zettlemoyer, Liang, Cand{\`e}s, and Hashimoto}]{muennighoff2025s1}
Niklas Muennighoff, Zitong Yang, Weijia Shi, Xiang~Lisa Li, Li~Fei-Fei, Hannaneh Hajishirzi, Luke Zettlemoyer, Percy Liang, Emmanuel Cand{\`e}s, and Tatsunori Hashimoto. 2025.
\newblock s1: Simple test-time scaling.
\newblock \emph{arXiv preprint arXiv:2501.19393}.

\bibitem[{Nahid and Rafiei(2024)}]{nahid2024tabsqlify}
Md~Mahadi~Hasan Nahid and Davood Rafiei. 2024.
\newblock Tabsqlify: Enhancing reasoning capabilities of llms through table decomposition.
\newblock \emph{arXiv preprint arXiv:2404.10150}.

\bibitem[{Nan et~al.(2022)Nan, Hsieh, Mao, Lin, Verma, Zhang, Kry{\'s}ci{\'n}ski, Schoelkopf, Kong, Tang, Mutuma, Rosand, Trindade, Bandaru, Cunningham, Xiong, Radev, and Radev}]{nan-etal-2022-fetaqa}
Linyong Nan, Chiachun Hsieh, Ziming Mao, Xi~Victoria Lin, Neha Verma, Rui Zhang, Wojciech Kry{\'s}ci{\'n}ski, Hailey Schoelkopf, Riley Kong, Xiangru Tang, Mutethia Mutuma, Ben Rosand, Isabel Trindade, Renusree Bandaru, Jacob Cunningham, Caiming Xiong, Dragomir Radev, and Dragomir Radev. 2022.
\newblock \href {https://doi.org/10.1162/tacl_a_00446} {{F}e{T}a{QA}: Free-form table question answering}.
\newblock \emph{Transactions of the Association for Computational Linguistics}, 10:35--49.

\bibitem[{Nan et~al.(2024)Nan, Zhang, Zou, Zhao, Zhou, and Cohan}]{nan-etal-2024-evaluating}
Linyong Nan, Ellen Zhang, Weijin Zou, Yilun Zhao, Wenfei Zhou, and Arman Cohan. 2024.
\newblock \href {https://doi.org/10.18653/v1/2024.findings-naacl.284} {On evaluating the integration of reasoning and action in {LLM} agents with database question answering}.
\newblock In \emph{Findings of the Association for Computational Linguistics: NAACL 2024}, pages 4556--4579, Mexico City, Mexico. Association for Computational Linguistics.

\bibitem[{Newman et~al.(2024)Newman, Lee, Naik, Siangliulue, Fok, Kim, Weld, Chang, and Lo}]{newman2024arxivdigestables}
Benjamin Newman, Yoonjoo Lee, Aakanksha Naik, Pao Siangliulue, Raymond Fok, Juho Kim, Daniel~S Weld, Joseph~Chee Chang, and Kyle Lo. 2024.
\newblock Arxivdigestables: Synthesizing scientific literature into tables using language models.
\newblock \emph{arXiv preprint arXiv:2410.22360}.

\bibitem[{OpenAI(2024)}]{openai2024gpt4o}
OpenAI. 2024.
\newblock \href {https://arxiv.org/abs/2410.21276} {Gpt-4o system card}.
\newblock \emph{arXiv preprint arXiv:2410.21276}.

\bibitem[{Ouyang et~al.(2025)Ouyang, Yan, Luo, Cheng, Liu, Liu, Yu, and Wang}]{Agent-R1}
Jie Ouyang, Ruiran Yan, Yucong Luo, Mingyue Cheng, Qi~Liu, Zirui Liu, Shuo Yu, and Daoyu Wang. 2025.
\newblock \href {https://github.com/0russwest0/Agent-R1} {Training powerful llm agents with end-to-end reinforcement learning}.

\bibitem[{Parikh et~al.(2020)Parikh, Wang, Gehrmann, Faruqui, Dhingra, Yang, and Das}]{parikh-etal-2020-totto}
Ankur Parikh, Xuezhi Wang, Sebastian Gehrmann, Manaal Faruqui, Bhuwan Dhingra, Diyi Yang, and Dipanjan Das. 2020.
\newblock \href {https://doi.org/10.18653/v1/2020.emnlp-main.89} {{ToTTo}: A controlled table-to-text generation dataset}.
\newblock In \emph{Proceedings of the 2020 Conference on Empirical Methods in Natural Language Processing (EMNLP)}, pages 1173--1186, Online. Association for Computational Linguistics.

\bibitem[{Pasupat and Liang(2015)}]{pasupat-liang-2015-compositional}
Panupong Pasupat and Percy Liang. 2015.
\newblock \href {https://doi.org/10.3115/v1/P15-1142} {Compositional semantic parsing on semi-structured tables}.
\newblock In \emph{Proceedings of the 53rd Annual Meeting of the Association for Computational Linguistics and the 7th International Joint Conference on Natural Language Processing (Volume 1: Long Papers)}, pages 1470--1480, Beijing, China. Association for Computational Linguistics.

\bibitem[{Pfister and Jud(2025)}]{pfister2025understanding}
Rolf Pfister and Hansueli Jud. 2025.
\newblock Understanding and benchmarking artificial intelligence: Openai's o3 is not agi.
\newblock \emph{arXiv preprint arXiv:2501.07458}.

\bibitem[{Qi et~al.(2024)Qi, Ma, Xu, Zhang, Yang, and Yang}]{qi2024mutual}
Zhenting Qi, Mingyuan Ma, Jiahang Xu, Li~Lyna Zhang, Fan Yang, and Mao Yang. 2024.
\newblock Mutual reasoning makes smaller llms stronger problem-solvers.
\newblock \emph{arXiv preprint arXiv:2408.06195}.

\bibitem[{Shao et~al.(2024)Shao, Wang, Zhu, Xu, Song, Bi, Zhang, Zhang, Li, Wu, and Guo}]{shao2024deepseekmath}
Zhihong Shao, Peiyi Wang, Qihao Zhu, Runxin Xu, Junxiao Song, Xiao Bi, Haowei Zhang, Mingchuan Zhang, Y.~K. Li, Y.~Wu, and Daya Guo. 2024.
\newblock \href {https://arxiv.org/abs/2402.03300} {Deepseekmath: Pushing the limits of mathematical reasoning in open language models}.
\newblock \emph{Preprint}, arXiv:2402.03300.

\bibitem[{Su et~al.(2024)Su, Wang, Ye, Zhou, Zhang, Chen, Zhu, Wang, Xu, Chen, Li, Lan, Tian, Yuan, Zhao, Zhou, Shou, Zha, Long, Li, Wu, Zhang, Huang, Yang, Zhang, Ye, Zhu, Hu, Gu, Sun, Li, Yang, and Xiao}]{su2024tablegpt2}
Aofeng Su, Aowen Wang, Chao Ye, Chen Zhou, Ga~Zhang, Gang Chen, Guangcheng Zhu, Haobo Wang, Haokai Xu, Hao Chen, Haoze Li, Haoxuan Lan, Jiaming Tian, Jing Yuan, Junbo Zhao, Junlin Zhou, Kaizhe Shou, Liangyu Zha, Lin Long, and 14 others. 2024.
\newblock \href {https://arxiv.org/abs/2411.02059} {Tablegpt2: A large multimodal model with tabular data integration}.
\newblock \emph{Preprint}, arXiv:2411.02059.

\bibitem[{Team et~al.(2025)Team, Du, Gao, Xing, Jiang, Chen, Li, Xiao, Du, Liao et~al.}]{team2025kimi}
Kimi Team, Angang Du, Bofei Gao, Bowei Xing, Changjiu Jiang, Cheng Chen, Cheng Li, Chenjun Xiao, Chenzhuang Du, Chonghua Liao, and 1 others. 2025.
\newblock Kimi k1. 5: Scaling reinforcement learning with llms.
\newblock \emph{arXiv preprint arXiv:2501.12599}.

\bibitem[{Team(2024)}]{qwen2.5}
Qwen Team. 2024.
\newblock \href {https://qwenlm.github.io/blog/qwen2.5/} {Qwen2.5: A party of foundation models}.

\bibitem[{Team(2025)}]{qwq32b}
Qwen Team. 2025.
\newblock \href {https://qwenlm.github.io/blog/qwq-32b/} {Qwq-32b: Embracing the power of reinforcement learning}.

\bibitem[{Touvron et~al.(2023)Touvron, Lavril, Izacard, Martinet, Lachaux, Lacroix, Rozi{\`e}re, Goyal, Hambro, Azhar et~al.}]{touvron2023llama}
Hugo Touvron, Thibaut Lavril, Gautier Izacard, Xavier Martinet, Marie-Anne Lachaux, Timoth{\'e}e Lacroix, Baptiste Rozi{\`e}re, Naman Goyal, Eric Hambro, Faisal Azhar, and 1 others. 2023.
\newblock Llama: Open and efficient foundation language models.
\newblock \emph{arXiv preprint arXiv:2302.13971}.

\bibitem[{Wang et~al.(2025)Wang, Wang, Wang, Zhang, Li, Yang, Yu, Nguyen, Liu, Gottlieb et~al.}]{wang2025ragen}
Zihan Wang, Kangrui Wang, Qineng Wang, Pingyue Zhang, Linjie Li, Zhengyuan Yang, Kefan Yu, Minh~Nhat Nguyen, Licheng Liu, Eli Gottlieb, and 1 others. 2025.
\newblock Ragen: Understanding self-evolution in llm agents via multi-turn reinforcement learning.
\newblock \emph{arXiv preprint arXiv:2504.20073}.

\bibitem[{Weller et~al.(2025)Weller, Ricci, Yang, Yates, Lawrie, and Durme}]{weller2025rank1}
Orion Weller, Kathryn Ricci, Eugene Yang, Andrew Yates, Dawn Lawrie, and Benjamin~Van Durme. 2025.
\newblock \href {https://arxiv.org/abs/2502.18418} {Rank1: Test-time compute for reranking in information retrieval}.
\newblock \emph{Preprint}, arXiv:2502.18418.

\bibitem[{Wen et~al.(2025)Wen, Cai, Xiao, He, An, Duan, Du, Liu, Tang, Lv et~al.}]{wen2025light}
Liang Wen, Yunke Cai, Fenrui Xiao, Xin He, Qi~An, Zhenyu Duan, Yimin Du, Junchen Liu, Lifu Tang, Xiaowei Lv, and 1 others. 2025.
\newblock Light-r1: Curriculum sft, dpo and rl for long cot from scratch and beyond.
\newblock \emph{arXiv preprint arXiv:2503.10460}.

\bibitem[{Wiseman et~al.(2017)Wiseman, Shieber, and Rush}]{wiseman-etal-2017-challenges}
Sam Wiseman, Stuart Shieber, and Alexander Rush. 2017.
\newblock \href {https://doi.org/10.18653/v1/D17-1239} {Challenges in data-to-document generation}.
\newblock In \emph{Proceedings of the 2017 Conference on Empirical Methods in Natural Language Processing}, pages 2253--2263, Copenhagen, Denmark. Association for Computational Linguistics.

\bibitem[{Wu et~al.(2025)Wu, Yang, Chai, Zhang, Liu, Du, Liang, Shu, Cheng, Sun et~al.}]{wu2025tablebench}
Xianjie Wu, Jian Yang, Linzheng Chai, Ge~Zhang, Jiaheng Liu, Xeron Du, Di~Liang, Daixin Shu, Xianfu Cheng, Tianzhen Sun, and 1 others. 2025.
\newblock Tablebench: A comprehensive and complex benchmark for table question answering.
\newblock In \emph{Proceedings of the AAAI Conference on Artificial Intelligence}, volume~39, pages 25497--25506.

\bibitem[{Xia et~al.(2025)Xia, Shen, Zhu, Zhang, Wang, Zhang, Liu, Xiao, Dong, Zhao et~al.}]{xia2025mimo}
Bingquan Xia, Bowen Shen, Dawei Zhu, Di~Zhang, Gang Wang, Hailin Zhang, Huaqiu Liu, Jiebao Xiao, Jinhao Dong, Liang Zhao, and 1 others. 2025.
\newblock Mimo: Unlocking the reasoning potential of language model--from pretraining to posttraining.
\newblock \emph{arXiv preprint arXiv:2505.07608}.

\bibitem[{Xia and Luo(2025)}]{xia2025gui}
Xiaobo Xia and Run Luo. 2025.
\newblock Gui-r1: A generalist r1-style vision-language action model for gui agents.
\newblock \emph{arXiv preprint arXiv:2504.10458}.

\bibitem[{Xie et~al.(2025)Xie, Gao, Ren, Luo, Hong, Dai, Zhou, Qiu, Wu, and Luo}]{xie2025logic}
Tian Xie, Zitian Gao, Qingnan Ren, Haoming Luo, Yuqian Hong, Bryan Dai, Joey Zhou, Kai Qiu, Zhirong Wu, and Chong Luo. 2025.
\newblock Logic-rl: Unleashing llm reasoning with rule-based reinforcement learning.
\newblock \emph{arXiv preprint arXiv:2502.14768}.

\bibitem[{Xu et~al.(2025)Xu, Jin, Li, Song, Sun, and Yuan}]{xu2025llavacot}
Guowei Xu, Peng Jin, Hao Li, Yibing Song, Lichao Sun, and Li~Yuan. 2025.
\newblock \href {https://arxiv.org/abs/2411.10440} {Llava-cot: Let vision language models reason step-by-step}.
\newblock \emph{Preprint}, arXiv:2411.10440.

\bibitem[{Ye et~al.(2025)Ye, Huang, Xiao, Chern, Xia, and Liu}]{ye2025limo}
Yixin Ye, Zhen Huang, Yang Xiao, Ethan Chern, Shijie Xia, and Pengfei Liu. 2025.
\newblock Limo: Less is more for reasoning.
\newblock \emph{arXiv preprint arXiv:2502.03387}.

\bibitem[{Ye et~al.(2023)Ye, Hui, Yang, Li, Huang, and Li}]{10.1145/3539618.3591708}
Yunhu Ye, Binyuan Hui, Min Yang, Binhua Li, Fei Huang, and Yongbin Li. 2023.
\newblock \href {https://doi.org/10.1145/3539618.3591708} {Large language models are versatile decomposers: Decomposing evidence and questions for table-based reasoning}.
\newblock In \emph{Proceedings of the 46th International ACM SIGIR Conference on Research and Development in Information Retrieval}, SIGIR '23, page 174–184, New York, NY, USA. Association for Computing Machinery.

\bibitem[{Yu et~al.(2025{\natexlab{a}})Yu, Chen, and Wang}]{yu2025tablecriticmultiagentframeworkcollaborative}
Peiying Yu, Guoxin Chen, and Jingjing Wang. 2025{\natexlab{a}}.
\newblock \href {https://arxiv.org/abs/2502.11799} {Table-critic: A multi-agent framework for collaborative criticism and refinement in table reasoning}.
\newblock \emph{Preprint}, arXiv:2502.11799.

\bibitem[{Yu et~al.(2025{\natexlab{b}})Yu, Zhang, Zhu, Yuan, Zuo, Yue, Fan, Liu, Liu, Liu et~al.}]{yu2025dapo}
Qiying Yu, Zheng Zhang, Ruofei Zhu, Yufeng Yuan, Xiaochen Zuo, Yu~Yue, Tiantian Fan, Gaohong Liu, Lingjun Liu, Xin Liu, and 1 others. 2025{\natexlab{b}}.
\newblock Dapo: An open-source llm reinforcement learning system at scale.
\newblock \emph{arXiv preprint arXiv:2503.14476}.

\bibitem[{Yuan et~al.(2024)Yuan, Li, Chen, Cui, Ding, Zhang, Zhou, Liu, and Peng}]{yuan2024free}
Lifan Yuan, Wendi Li, Huayu Chen, Ganqu Cui, Ning Ding, Kaiyan Zhang, Bowen Zhou, Zhiyuan Liu, and Hao Peng. 2024.
\newblock Free process rewards without process labels.
\newblock \emph{arXiv preprint arXiv:2412.01981}.

\bibitem[{Yue et~al.(2025)Yue, Chen, Lu, Zhao, Wang, Song, and Huang}]{yue2025does}
Yang Yue, Zhiqi Chen, Rui Lu, Andrew Zhao, Zhaokai Wang, Shiji Song, and Gao Huang. 2025.
\newblock Does reinforcement learning really incentivize reasoning capacity in llms beyond the base model?
\newblock \emph{arXiv preprint arXiv:2504.13837}.

\bibitem[{Zha et~al.(2023)Zha, Zhou, Li, Wang, Huang, Yang, Yuan, Su, Li, Su, Zhang, Zhou, Shou, Wang, Zhu, Lu, Ye, Ye, Ye, Zhang, Deng, Xu, Wang, Chen, and Zhao}]{zha2023tablegpt}
Liangyu Zha, Junlin Zhou, Liyao Li, Rui Wang, Qingyi Huang, Saisai Yang, Jing Yuan, Changbao Su, Xiang Li, Aofeng Su, Tao Zhang, Chen Zhou, Kaizhe Shou, Miao Wang, Wufang Zhu, Guoshan Lu, Chao Ye, Yali Ye, Wentao Ye, and 6 others. 2023.
\newblock \href {https://arxiv.org/abs/2307.08674} {Tablegpt: Towards unifying tables, nature language and commands into one gpt}.
\newblock \emph{Preprint}, arXiv:2307.08674.

\bibitem[{Zhang et~al.(2023)Zhang, Yue, Li, and Sun}]{zhang2023tablellama}
Tianshu Zhang, Xiang Yue, Yifei Li, and Huan Sun. 2023.
\newblock Tablellama: Towards open large generalist models for tables.
\newblock \emph{arXiv preprint arXiv:2311.09206}.

\bibitem[{Zhang et~al.(2024{\natexlab{a}})Zhang, Yue, Li, and Sun}]{zhang-etal-2024-tablellama}
Tianshu Zhang, Xiang Yue, Yifei Li, and Huan Sun. 2024{\natexlab{a}}.
\newblock \href {https://doi.org/10.18653/v1/2024.naacl-long.335} {{T}able{L}lama: Towards open large generalist models for tables}.
\newblock In \emph{Proceedings of the 2024 Conference of the North American Chapter of the Association for Computational Linguistics: Human Language Technologies (Volume 1: Long Papers)}, pages 6024--6044, Mexico City, Mexico. Association for Computational Linguistics.

\bibitem[{Zhang et~al.(2025{\natexlab{a}})Zhang, Luo, Zhang, Ma, Zhang, Li, Li, Yao, Xu, Zhou, Zhang-Li, Yu, Zhao, Li, and Tang}]{zhang2025tablellm}
Xiaokang Zhang, Sijia Luo, Bohan Zhang, Zeyao Ma, Jing Zhang, Yang Li, Guanlin Li, Zijun Yao, Kangli Xu, Jinchang Zhou, Daniel Zhang-Li, Jifan Yu, Shu Zhao, Juanzi Li, and Jie Tang. 2025{\natexlab{a}}.
\newblock \href {https://arxiv.org/abs/2403.19318} {Tablellm: Enabling tabular data manipulation by llms in real office usage scenarios}.
\newblock \emph{Preprint}, arXiv:2403.19318.

\bibitem[{Zhang et~al.(2024{\natexlab{b}})Zhang, Luo, Zhang, Ma, Zhang, Li, Li, Yao, Xu, Zhou et~al.}]{zhang2024tablellm}
Xiaokang Zhang, Sijia Luo, Bohan Zhang, Zeyao Ma, Jing Zhang, Yang Li, Guanlin Li, Zijun Yao, Kangli Xu, Jinchang Zhou, and 1 others. 2024{\natexlab{b}}.
\newblock Tablellm: Enabling tabular data manipulation by llms in real office usage scenarios.
\newblock \emph{arXiv preprint arXiv:2403.19318}.

\bibitem[{Zhang et~al.(2025{\natexlab{b}})Zhang, Wang, Dou, Zhu, and Che}]{zhang2025survey}
Xuanliang Zhang, Dingzirui Wang, Longxu Dou, Qingfu Zhu, and Wanxiang Che. 2025{\natexlab{b}}.
\newblock A survey of table reasoning with large language models.
\newblock \emph{Frontiers of Computer Science}, 19(9):199348.

\bibitem[{Zhao et~al.(2024{\natexlab{a}})Zhao, Chen, Cohan, and Zhao}]{zhao-etal-2024-tapera}
Yilun Zhao, Lyuhao Chen, Arman Cohan, and Chen Zhao. 2024{\natexlab{a}}.
\newblock \href {https://doi.org/10.18653/v1/2024.acl-long.692} {{T}a{PERA}: Enhancing faithfulness and interpretability in long-form table {QA} by content planning and execution-based reasoning}.
\newblock In \emph{Proceedings of the 62nd Annual Meeting of the Association for Computational Linguistics (Volume 1: Long Papers)}, pages 12824--12840, Bangkok, Thailand. Association for Computational Linguistics.

\bibitem[{Zhao et~al.(2024{\natexlab{b}})Zhao, Liu, Long, Zhang, Zhao, and Cohan}]{zhao-etal-2024-knowledgefmath}
Yilun Zhao, Hongjun Liu, Yitao Long, Rui Zhang, Chen Zhao, and Arman Cohan. 2024{\natexlab{b}}.
\newblock \href {https://doi.org/10.18653/v1/2024.acl-long.693} {Financemath: Knowledge-intensive math reasoning in finance domains}.
\newblock In \emph{Proceedings of the 62nd Annual Meeting of the Association for Computational Linguistics (Volume 1: Long Papers)}, pages 12841--12858, Bangkok, Thailand. Association for Computational Linguistics.

\bibitem[{Zhao et~al.(2024{\natexlab{c}})Zhao, Long, Liu, Kamoi, Nan, Chen, Liu, Tang, Zhang, and Cohan}]{zhao-etal-2024-docmath}
Yilun Zhao, Yitao Long, Hongjun Liu, Ryo Kamoi, Linyong Nan, Lyuhao Chen, Yixin Liu, Xiangru Tang, Rui Zhang, and Arman Cohan. 2024{\natexlab{c}}.
\newblock \href {https://doi.org/10.18653/v1/2024.acl-long.852} {{D}oc{M}ath-eval: Evaluating math reasoning capabilities of {LLM}s in understanding long and specialized documents}.
\newblock In \emph{Proceedings of the 62nd Annual Meeting of the Association for Computational Linguistics (Volume 1: Long Papers)}, pages 16103--16120, Bangkok, Thailand. Association for Computational Linguistics.

\bibitem[{Zhao et~al.(2022)Zhao, Nan, Qi, Zhang, and Radev}]{zhao-etal-2022-reastap}
Yilun Zhao, Linyong Nan, Zhenting Qi, Rui Zhang, and Dragomir Radev. 2022.
\newblock \href {https://doi.org/10.18653/v1/2022.emnlp-main.615} {{R}eas{TAP}: Injecting table reasoning skills during pre-training via synthetic reasoning examples}.
\newblock In \emph{Proceedings of the 2022 Conference on Empirical Methods in Natural Language Processing}, pages 9006--9018, Abu Dhabi, United Arab Emirates. Association for Computational Linguistics.

\bibitem[{Zhao et~al.(2023{\natexlab{a}})Zhao, Qi, Nan, Flores, and Radev}]{zhao-etal-2023-loft}
Yilun Zhao, Zhenting Qi, Linyong Nan, Lorenzo~Jaime Flores, and Dragomir Radev. 2023{\natexlab{a}}.
\newblock \href {https://doi.org/10.18653/v1/2023.eacl-main.40} {{L}o{FT}: Enhancing faithfulness and diversity for table-to-text generation via logic form control}.
\newblock In \emph{Proceedings of the 17th Conference of the European Chapter of the Association for Computational Linguistics}, pages 554--561, Dubrovnik, Croatia. Association for Computational Linguistics.

\bibitem[{Zhao et~al.(2023{\natexlab{b}})Zhao, Qi, Nan, Mi, Liu, Zou, Han, Chen, Tang, Xu, Radev, and Cohan}]{zhao-etal-2023-qtsumm}
Yilun Zhao, Zhenting Qi, Linyong Nan, Boyu Mi, Yixin Liu, Weijin Zou, Simeng Han, Ruizhe Chen, Xiangru Tang, Yumo Xu, Dragomir Radev, and Arman Cohan. 2023{\natexlab{b}}.
\newblock \href {https://doi.org/10.18653/v1/2023.emnlp-main.74} {{QTS}umm: Query-focused summarization over tabular data}.
\newblock In \emph{Proceedings of the 2023 Conference on Empirical Methods in Natural Language Processing}, pages 1157--1172, Singapore. Association for Computational Linguistics.

\bibitem[{Zhao et~al.(2023{\natexlab{c}})Zhao, Zhang, Si, Nan, Tang, and Cohan}]{zhao-etal-2023-investigating}
Yilun Zhao, Haowei Zhang, Shengyun Si, Linyong Nan, Xiangru Tang, and Arman Cohan. 2023{\natexlab{c}}.
\newblock \href {https://doi.org/10.18653/v1/2023.emnlp-industry.17} {Investigating table-to-text generation capabilities of large language models in real-world information seeking scenarios}.
\newblock In \emph{Proceedings of the 2023 Conference on Empirical Methods in Natural Language Processing: Industry Track}, pages 160--175, Singapore. Association for Computational Linguistics.

\bibitem[{Zhao et~al.(2023{\natexlab{d}})Zhao, Zhao, Nan, Qi, Zhang, Tang, Mi, and Radev}]{zhao-etal-2023-robut}
Yilun Zhao, Chen Zhao, Linyong Nan, Zhenting Qi, Wenlin Zhang, Xiangru Tang, Boyu Mi, and Dragomir Radev. 2023{\natexlab{d}}.
\newblock \href {https://doi.org/10.18653/v1/2023.acl-long.334} {{R}obu{T}: A systematic study of table {QA} robustness against human-annotated adversarial perturbations}.
\newblock In \emph{Proceedings of the 61st Annual Meeting of the Association for Computational Linguistics (Volume 1: Long Papers)}, pages 6064--6081, Toronto, Canada. Association for Computational Linguistics.

\bibitem[{Zhuang et~al.(2025)Zhuang, Ma, Koopman, Lin, and Zuccon}]{zhuang2025rankr1}
Shengyao Zhuang, Xueguang Ma, Bevan Koopman, Jimmy Lin, and Guido Zuccon. 2025.
\newblock \href {https://arxiv.org/abs/2503.06034} {Rank-r1: Enhancing reasoning in llm-based document rerankers via reinforcement learning}.
\newblock \emph{Preprint}, arXiv:2503.06034.

\end{thebibliography}

\newpage
\appendix

\clearpage
\onecolumn
\section{Dataset Details}
\subsection{Detailed Explanation of Training Data}
\label{sec:appendix-train-dataset}
Our training data originates from four benchmark datasets: WTQ, HiTab, TabFact, and FeTaQA.

\textbf{Training Data for RLVR.} For WTQ, HiTab, and FeTaQA, we used their complete training sets. For TabFact, which has a very large training set ($ > 90,000$ samples), we implemented a difficulty-based filtering process to create a more focused and challenging subset. Specifically, we used Qwen2.5-7B-Instruct to generate $16$ candidate responses for each sample. We retained all samples with a pass rate of $\leq 7 / 16$ and randomly sampled $10,000$ instances from those with a pass rate between $8/16$ and $15/16$. This resulted in a final set of $20,740$ samples for TabFact. After constructing the prompts, we filtered out any instances where the prompt length exceeded $4096$ tokens. This final, combined dataset ($48,563$ samples) was used for RLVR training.

\textbf{Training Data for SFT.} We used the RLVR dataset as a starting point. We then generated reasoning traces for each sample using DeepSeek-R1 and applied a stringent quality-control filter to retain only high-quality examples. For WTQ, HiTab, and TabFact, we kept only the samples where the generated answer was an exact match to the ground truth. For FeTaQA, we kept samples where the average of BLEU and ROUGE-L scores was $\geq 0.35$. This resulted in a high-quality SFT dataset of $33,601$ samples.

The data volumes at each stage are summarized below, as shown in Table~\ref{tab:data_volumes}.

\begin{table}[ht]
\centering
\small %
\begin{tabular}{lrrr}
\toprule
\textbf{Dataset} & \textbf{Raw Samples} & \textbf{Samples for RLVR} & \textbf{Samples for SFT (Filtered)} \\
\midrule
WTQ     & 14,152  & 13,706  & 9,524  \\
HiTab   & 7,301   & 6,793   & 4,170  \\
TabFact & 92,283  & 20,740  & 16,006 \\
FeTaQA  & 7,326   & 7,324   & 3,901  \\
\midrule
\textbf{Total} & \textbf{121,062} & \textbf{48,563} & \textbf{33,601} \\
\bottomrule
\end{tabular}
\caption{The data volumes at each stage.}
\label{tab:data_volumes}
\end{table}

Table~\ref{tab:sft_statistics} provides detailed statistics on the SFT data. Notably, the quality filtering significantly reduces the average response-to-prompt length ratio (from $1.73$ to $1.49$ overall). This aligns with the common observation that incorrect or lower-quality reasoning traces are often longer and more convoluted, reinforcing the effectiveness of our filtering strategy.

\begin{table}[ht]
\centering
\small %
\begin{tabular}{lrrrrrrc}
\toprule
\textbf{\makecell{Dataset / \\ Length (Chars)}} &
\textbf{\makecell{Min \\ Prompt}} &
\textbf{\makecell{Max \\ Prompt}} &
\textbf{\makecell{Avg. \\ Prompt}} &
\textbf{\makecell{Min \\ Resp.}} &
\textbf{\makecell{Max \\ Resp.}} &
\textbf{\makecell{Avg. \\ Resp.}} &
\textbf{\makecell{Avg. Resp/Prompt Ratio \\ (Before/After Filtering)}} \\
\midrule
WTQ & 625 & 14,192 & 2,114.67 & 465 & 28,023 & 2,356.26 & 1.78 / 1.45 \\
HiTab & 694 & 10,264 & 3,251.17 & 526 & 32,630 & 2,827.42 & 1.57 / 1.12 \\
TabFact & 624 & 8,906 & 1,767.64 & 529 & 28,762 & 2,474.04 & 1.76 / 1.61 \\
FeTaQA & 473 & 13,793 & 1,656.14 & 546 & 28,661 & 1,982.57 & 1.71 / 1.53 \\
\midrule
Overall & 473 & 14,192 & 2,037.17 & 465 & 32,630 & 2,427.45 & 1.73 / 1.49 \\
\bottomrule
\end{tabular}%
\caption{Detailed statistics on the SFT data.}
\label{tab:sft_statistics}
\end{table}

\clearpage

\subsection{Information of Evaluated Table Reasoning Datasets}
\label{sec:appendix-dataset}
\begin{table*}[ht]
\centering
\small
\resizebox{\textwidth}{!}{
\begin{tabular}{lllll}
\toprule
\textbf{Task Category} & \textbf{Task Name} & \textbf{Dataset} & \textbf{Task Description} & \textbf{Metric} \\
\midrule
\multirow{4}{*}{Training}
  & Table QA & WTQ, HiTab & QA over flat or hierarchical tables & Acc. \\
  & Free-form QA & FeTaQA & Generate long-form answers from tables & BLEU \\
  & Table Fact Verification & TabFact & Verify factual correctness based on a table & Acc. \\
\midrule
\multirow{4}{*}{In-Domain}
  & Table QA & WTQ, HiTab & In-domain table QA evaluation & Acc. \\
  & Table Fact Verification & TabFact & In-domain fact verification evaluation & Acc.\\
  & Free-form QA & FeTaQA & Evaluate long-form generation quality & BLEU  \\
\midrule
\multirow{4}{*}{Out-of-Domain}
  & Table QA & TabMCQ, TMWP, FinQA & Multiple-choice and word problem QA & Acc. \\
  & Numerical Reasoning QA & FinQA, TABMWP & Financial and numerical table QA & Acc. \\
  & Table Fact Verification & InfoTabs, PHT, Feverous & Health and factual judgment from tables & Acc. \\
  & Free-form QA & ToTTo, Qtsumm, R.W. & Table-to-text generation from highlights & BLEU/R-L \\
\bottomrule
\end{tabular}
}
\caption{Overview of datasets used in training and evaluation.}
\label{tab:our_dataset}
\end{table*}

\section{Experiment Setup}\label{app:exp-setup}
\subsection{Baseline Systems}
We benchmark our approach against a comprehensive set of strong baselines, encompassing both proprietary and open-source models. Among proprietary models, we include GPT-4.1 and GPT-4.1 mini. For open-source baselines, we evaluate general purpose LLMs from the Qwen2.5 and LLaMA3 series, reasoning models such as Deepseek-R1, its official distilled variant, our SFT model distilled from Deepseek-R1 on table reasoning data (as described in Section~3.2), and QwQ-32B, as well as table-oriented models including TableLlama, TableLLM, and TableBenchLLM.

\subsection{Inference Setup}
For all open-source models, inference is performed using the vLLM framework, while for closed-source models, the official OpenAI API is utilized. The maximum output length is set to 2048 tokens for most models. However, for reasoning models, this limit is increased to 18,000 tokens to accommodate their long chain-of-thought generation. The temperature is set to 0.6 and the top-p value to 0.95. All inference processes are conducted on four NVIDIA A100-80G GPUs.

\subsection{Evaluated Model Configuration}
\begin{table*}[h]
    \centering
    \small
    \begin{tabular}{lll}
        \toprule
        \textbf{Model} & \textbf{Citation} & \textbf{Version} \\ \midrule

        GPT-4.1 & \citet{openai2024gpt4o} & gpt-4.1-2025-04-14 \\
        GPT-4.1 mini & \citet{openai2024gpt4o} & gpt-4.1-mini-2025-04-14 \\ \midrule

        Qwen2.5-7B & \citet{qwen2.5} & \href{https://huggingface.co/Qwen/Qwen2.5-7B-Instruct}{Qwen/Qwen2.5-7B-Instruct} \\
        Qwen2.5-14B & \citet{qwen2.5} & \href{https://huggingface.co/Qwen/Qwen2.5-14B-Instruct}{Qwen/Qwen2.5-14B-Instruct} \\
        Qwen2.5-32B & \citet{qwen2.5} & \href{https://huggingface.co/Qwen/Qwen2.5-32B-Instruct}{Qwen/Qwen2.5-32B-Instruct} \\
        QwQ-32B & \citet{qwq32b} & \href{https://huggingface.co/Qwen/QwQ-32B}{Qwen/QwQ-32B} \\

        Llama-3.1-8B & \citet{touvron2023llama} & \href{https://huggingface.co/meta-llama/Llama-3.1-8B-Instruct}{meta-llama/Llama-3.1-8B-Instruct} \\ \midrule

        DeepSeek-R1-7B & \citet{guo2025deepseek} & \href{https://huggingface.co/deepseek-ai/DeepSeek-R1-Distill-Qwen-7B}{deepseek-ai/DeepSeek-R1-Distill-Qwen-7B} \\
        DeepSeek-R1-14B & \citet{guo2025deepseek} & \href{https://huggingface.co/deepseek-ai/DeepSeek-R1-Distill-Qwen-14B}{deepseek-ai/DeepSeek-R1-Distill-Qwen-14B} \\
        DeepSeek-R1-32B & \citet{guo2025deepseek} & \href{https://huggingface.co/deepseek-ai/DeepSeek-R1-Distill-Qwen-32B}{deepseek-ai/DeepSeek-R1-Distill-Qwen-32B} \\
        DeepSeek-R1 & \citet{guo2025deepseek} & \href{hhttps://huggingface.co/deepseek-ai/DeepSeek-R1}{deepseek-ai/DeepSeek-R1} \\ \midrule
        TableGPT2-7B & \citet{su2024tablegpt2} & \href{https://huggingface.co/tablegpt/TableGPT2-7B}{TableGPT/TableGPT2-7B} \\
        TableLLM-13B & \citet{zhang2024tablellm} & \href{https://huggingface.co/RUCKBReasoning/TableLLM-13b}{RUCKBReasoning/TableLLM-13B} \\ 
        TableLlama-7B & \citet{zhang2023tablellama} & \href{https://huggingface.co/osunlp/TableLlama}{osunlp/TableLlama} \\ 
        TableBenchLLM & \citet{zhang2024tablellm} & \href{https://huggingface.co/Multilingual-Multimodal-NLP/TableLLM-Llama3.1-8B}{Multilingual-Multimodal-NLP/TableLLM-Llama3.1-8B} \\ 
        
        \bottomrule
    \end{tabular}
    \caption{Model List.}
    \label{tab:model_list}
\end{table*}

\section{Prompts}
\label{sec:appendix-prompt-template}
\begin{tcolorbox}[colback=black!3!white, colframe=black!70!white, title=System prompt used to guide structured response generation, fontupper=\footnotesize, fonttitle=\footnotesize]
A conversation between User and Assistant. The user asks a question, and the assistant solves it. The assistant first thinks about the reasoning process in the mind and then provides the user with the answer. The reasoning process and answer are enclosed within <think> </think> and <answer> </answer> tags, respectively, i.e., <think> reasoning process here </think> <answer> answer here </answer>.
\end{tcolorbox}

\begin{tcolorbox}[colback=black!3!white, colframe=black!70!white, title=Prompt Template for TQA, fontupper=\footnotesize, fonttitle=\footnotesize]
    \textbf{Instruction:} This is a short-answer table QA task. Answer the question based on the provided table. \\
    
    \textbf{Table} \\
    Table Title: \texttt{\{table\_title\}} \\
    Table Content: \texttt{\{table\_repr (markdown / html)\}} \\
    
    \textbf{Question:} \texttt{\{question\}} \\
    
    \textbf{Answer Format:} \\
    The final answer should be concise and use the following format:
    \begin{verbatim}
```json
{
    "answer": ["answer1", "answer2", ...]
}
```
\end{verbatim}
\end{tcolorbox}

\begin{tcolorbox}[colback=black!3!white, colframe=black!70!white, title=Prompt Template for TFV, fontupper=\footnotesize, fonttitle=\footnotesize]
    \textbf{Instruction:} This is a table fact verification task. The goal is to determine whether the given statement is entailed or refuted by the table. \\

    \textbf{Table} \\
    Table Title: \texttt{\{table\_title\}} \\
    Table Content: \texttt{\{table\_repr (markdown / html)\}} \\

    \textbf{Statement:} \texttt{\{statement\}} \\

    \textbf{Answer Format:} \\
    The final answer should be either "entailed" or "refuted" and use the following format:
\begin{verbatim}
```json
{
    "answer": "entailed" or "refuted"
}
```
\end{verbatim}
\end{tcolorbox}

\begin{tcolorbox}[colback=black!3!white, colframe=black!70!white, title=Prompt Template for Free-Form TQA, fontupper=\footnotesize, fonttitle=\footnotesize]
    \textbf{Instruction:} This is a free-form table QA task. Answer the question based on the provided table. \\

    \textbf{Table} \\
    Table Title: \texttt{\{table\_title\}} \\
    Table Content: \texttt{\{table\_repr (markdown / html)\}} \\

    \textbf{Question:} \texttt{\{question\}} \\

    \textbf{Answer Format:} \\
    The final answer should be a sentence and use the following format:
\begin{verbatim}
```json
{
    "answer": "your_generated_sentence_here"
}
```
\end{verbatim}
\end{tcolorbox}

\begin{tcolorbox}[colback=black!3!white, colframe=black!70!white, title=Prompt Template for LLM-as-a-Judge, fontupper=\footnotesize, fonttitle=\footnotesize]
    You are given two answers for a short-answer Table QA task: \textbf{response} and \textbf{ground\_truth}. \\

    - \textbf{response}: This is the LLM's answer to the task. It may include reasoning steps and a final answer. \\
    - \textbf{ground\_truth}: A list of short answers, typically 2-3 word noun phrases or numbers. \\

    Your task is to determine whether the response is fully correct, using these rules: \\
    - \textbf{Noun phrases}: Considered correct if meaning matches ground\_truth regardless of wording. \\
    - \textbf{Numbers}: Considered correct if numerically close (tolerance < 0.01). \\
        - Every ground\_truth item must be matched in the response. Order doesn't matter. \\

Your output must be in the following format:
\begin{verbatim}
```json
{
    "judgement": "correct" or "incorrect"
}
```
\end{verbatim}
Do not provide any explanation or additional output.\\

\textbf{Input:} \\
\textbf{Response:} \{response\} \\
\textbf{Ground\_truth:} \{ground\_truth\} \\

Evaluate and output the judgement.
\end{tcolorbox}
\label{sec:appendix-prompt-llm-as-a-judge}

\section{Analysis}
\subsection{Qualitative Analysis Cases}\label{app:example}

\begin{figure*}[h]
  \includegraphics[width=\linewidth]{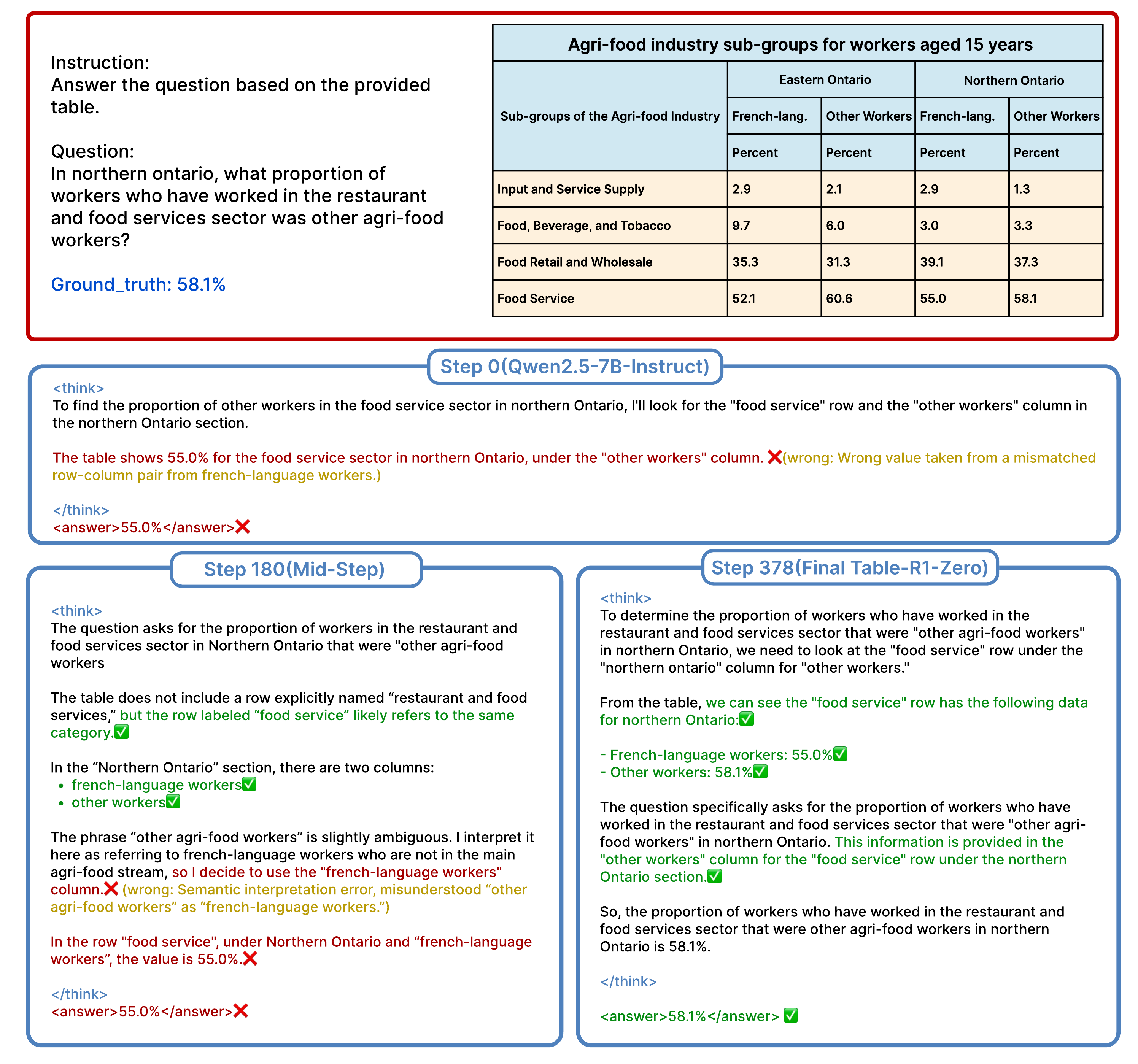}
  \caption{Qualitative Analysis Case Example for TQA tasks}
\end{figure*}

\begin{figure}[h]
  \includegraphics[width=\linewidth]{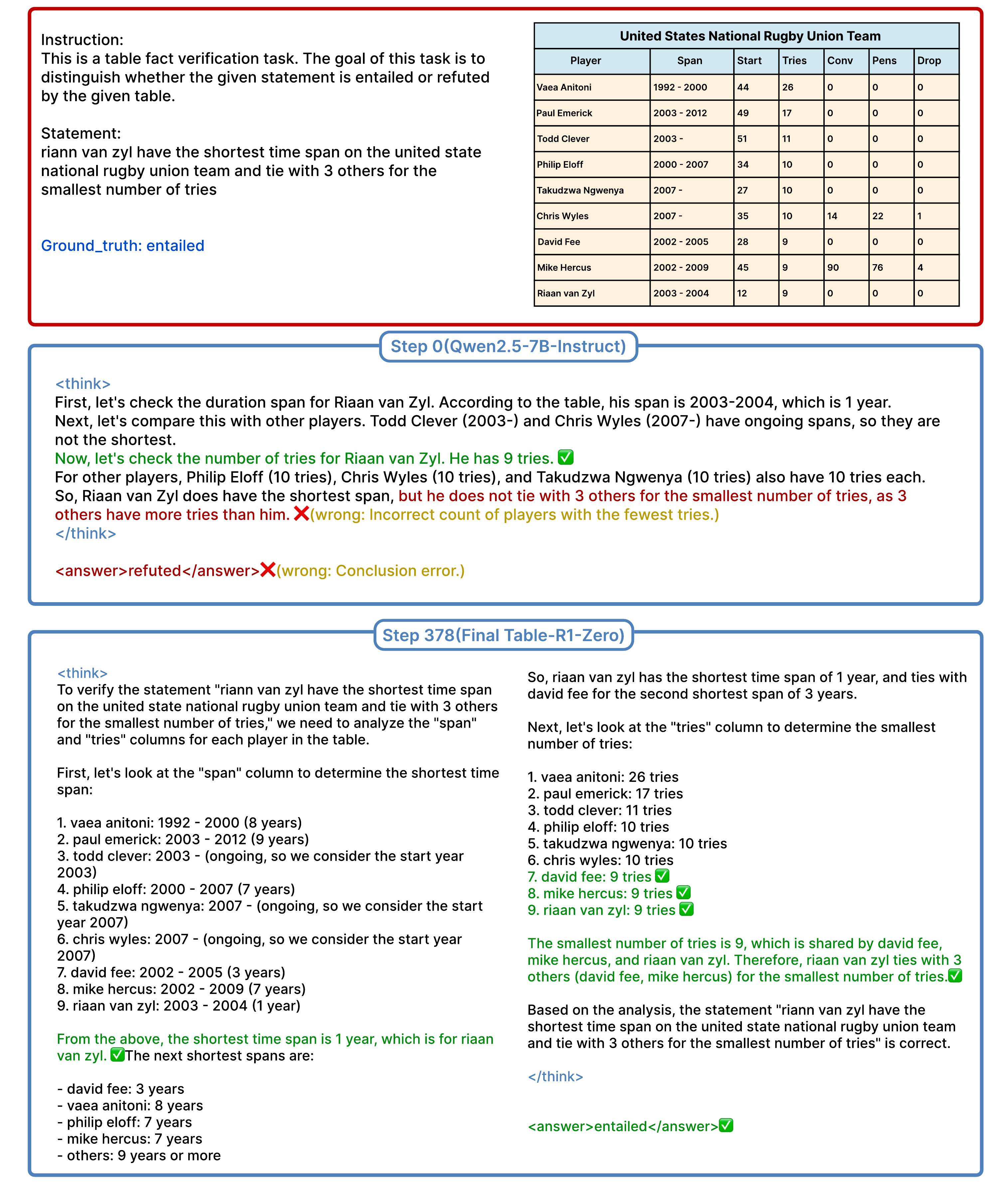}
  \caption{Qualitative Analysis Case Example for TFV tasks}
\end{figure}

\subsection{Reasoning Capacity Boundary}
\begin{figure}[!t]
\includegraphics[width=0.48\columnwidth]{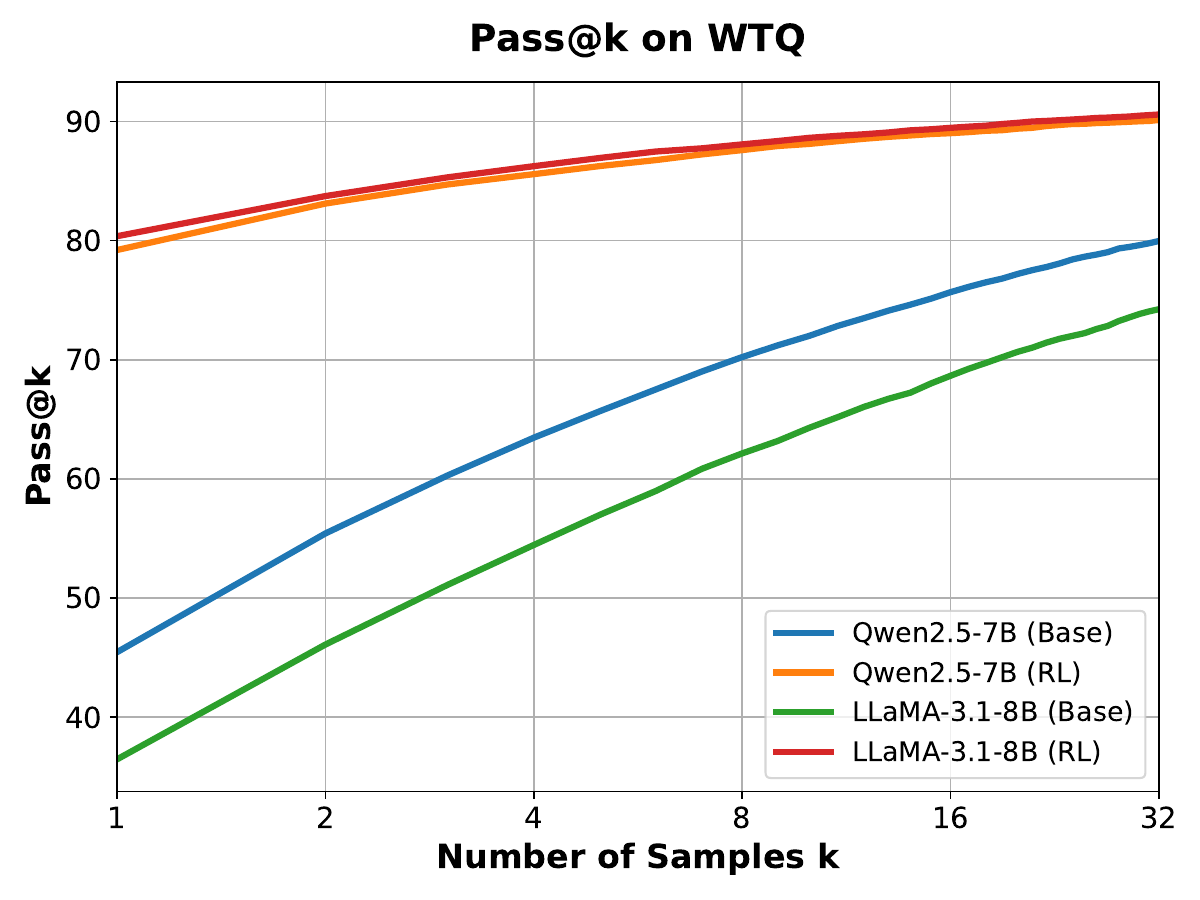}
  \includegraphics[width=0.48\columnwidth]{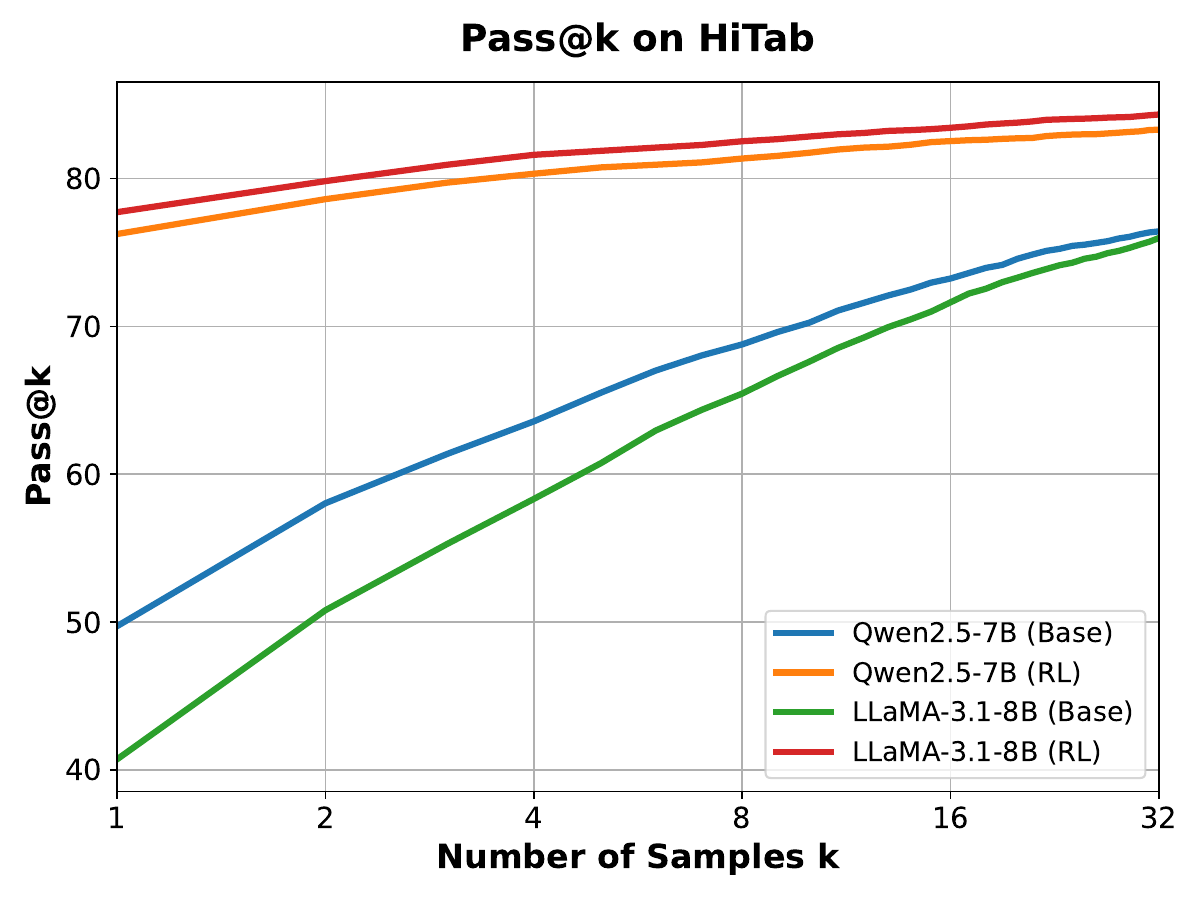}
  \caption {Pass@k performance on WTQ and HiTab.}
  \label{fig:passk}
\end{figure}

\subsection{Ablation Studies}
\begin{table*}[!t]
\label{tab:ablation-studies}
\centering
\setlength{\tabcolsep}{2pt}

\resizebox{\textwidth}{!}{
\begin{tabular}{l*{13}{c}}
\toprule
\multirow{3}{*}{\textbf{Model}} & \multicolumn{4}{c}{\textbf{In-domain Performance}} & \multicolumn{9}{c}{\textbf{Out-of-domain Performance}} \\
\cmidrule(lr){2-5} \cmidrule(lr){6-14}

& \multicolumn{1}{c}{\textbf{FF-TQA}} & \multicolumn{1}{c}{\textbf{TFV}} & \multicolumn{2}{c}{\textbf{TQA}} & \multicolumn{3}{c}{\textbf{FF-TQA}} & \multicolumn{3}{c}{\textbf{TFV}} & \multicolumn{3}{c}{\textbf{TQA}} \\
\cmidrule(lr){2-2} \cmidrule(lr){3-3} \cmidrule(lr){4-5} \cmidrule(lr){6-8} \cmidrule(lr){9-11} \cmidrule(lr){12-14}

& \multicolumn{1}{c}{FeTaQA} & \multicolumn{1}{c}{TabFact} & \multicolumn{1}{c}{WTQ} & \multicolumn{1}{c}{HiTab} & \multicolumn{1}{c}{ToTTo} & \multicolumn{1}{c}{QTSum} & \multicolumn{1}{c}{R.W.} & \multicolumn{1}{c}{InfoTabs} & \multicolumn{1}{c}{PHT} & \multicolumn{1}{c}{Feverous} & \multicolumn{1}{c}{TMCQ} & \multicolumn{1}{c}{TMWP} & \multicolumn{1}{c}{FinQA} \\
\midrule

\multicolumn{14}{c}{\textit{\textbf{Model-level Ablation}}} \\
\addlinespace[0.5em]

\multicolumn{14}{l}{\textit{Base vs. Instruct}} \\
\addlinespace[0.2em]
Qwen2.5-7B              & 18.2 & 67.7 & 51.0 & 55.2 & 9.4 & 37.8 & 16.4 & 74.5 & 75.0 & 74.1 & 78.7 & 80.3 & 58.4 \\
Table-R1-Qwen2.5-7B       & 29.8 & 87.3 & 78.5 & 75.2 & 15.9 & 38.1 & 18.2 & 87.9 & 91.6 & 78.4 & 84.8 & 94.4 & 61.2 \\
Qwen2.5-7B-Instruct         & 21.0 & 72.2 & 54.8 & 61.8 & 16.0 & 39.5 & 19.3 & 78.6 & 70.7 & 74.6 & 87.4 & 85.0 & 66.4 \\
Table-R1-Qwen2.5-7B-Instruct                 & 30.6 & 87.6 & 79.8 & 78.1 & 19.8 & 43.1 & 20.0 & 83.7 & 88.0 & 76.2 & 93.0 & 96.4 & 70.8 \\

\addlinespace[0.2em]
\cdashlinelr{1-14}
\addlinespace[0.5em]

\multicolumn{14}{l}{\textit{Model Architecture Comparison}} \\
\addlinespace[0.2em]
Llama-3.1-8B            & 10.2 & 21.2 & 35.0 & 44.7 & 6.8 & 18.6 & 10.9 & 21.6 & 26.8 & 19.4 & 68.9 & 43.1 & 21.9 \\
Table-R1-Llama-3.1-8B & 30.4 & 50.3 & 61.5 & 76.3 & 19.8 & 33.5 & 18.0 & 50.0 & 69.7 & 50.6 & 93.1 & 56.8 & 19.3 \\
Llama-3.1-8B-Instruct       & 21.7 & 74.1 & 52.3 & 58.2 & 16.5 & 31.6 & 18.1 & 84.1 & 82.5 & 78.3 & 49.5 & 72.0 & 57.1 \\
Table-R1-Llama-3.1-8B-Instruct    & 32.7 & 87.6 & 81.2 & 81.4 & 22.3 & 30.2 & 17.7 & 87.9 & 91.6 & 80.2 & 68.6 & 84.6 & 62.3 \\

\addlinespace[0.2em]
\cdashlinelr{1-14}
\addlinespace[0.5em]

\multicolumn{14}{l}{\textit{SFT on Domain-Specific Data}} \\
\addlinespace[0.2em]
DeepSeek-R1-7B     & 19.1 & 79.6 & 57.8 & 46.2 & 10.7 & 37.2 & 18.0 & 85.7 & 87.1 & 77.5 & 80.9 & 94.0 & 66.8 \\
Table-SFT-Qwen2.5-7B                 & 25.3 & 89.9 & 81.9 & 78.3 & 14.1 & 38.8 & 18.8 & 88.8 & 84.6 & 76.0 & 90.9 & 96.6 & 71.7 \\
Table-SFT-Llama-3.1-8B                 & 26.0 & 91.1 & 83.8 & 81.8 & 13.7 & 36.6 & 16.6 & 89.8 & 85.8 & 79.4 & 90.8 & 89.0 & 64.3 \\
\midrule

\multicolumn{14}{c}{\textit{\textbf{Task-level Ablation}}} \\
\addlinespace[0.5em]
\multicolumn{14}{l}{\textit{Cross-task Generalization}} \\
\addlinespace[0.2em]
Table-R1-TQA-Qwen2.5-7B           & 14.1 & 86.2 & 79.2 & 77.4 & 16.5 & 36.4 & 19.8 & 89.7 & 89.9 & 82.7 & 93.5 & 96.1 & 65.9 \\
Table-R1-TQA-Llama-3.1-8B         & 12.3 & 83.9 & 81.7 & 81.0 & 17.2 & 26.9 & 19.2 & 87.3 & 87.2 & 77.1 & 80.3 & 82.8 & 58.6 \\
Table-SFT-TQA-Qwen2.5-7B          & 13.3 & 87.7 & 79.8 & 77.0 & 13.4 & 27.6 & 18.1 & 88.5 & 87.1 & 78.0 & 90.1 & 95.7 & 69.0 \\
Table-SFT-TQA-Llama-3.1-8B & 9.6 & 88.8 & 82.4 & 80.5 & 10.3 & 20.6 & 12.6 & 89.1 & 90.4 & 82.1 & 89.9 & 86.9 & 62.9 \\
\midrule

\multicolumn{14}{c}{\textit{\textbf{Format Ablation}}} \\
\addlinespace[0.5em]
\multicolumn{14}{l}{\textit{Effect of Format Reward}} \\
\addlinespace[0.2em]
Table-R1-Explicit-Qwen2.5-7B    & 29.0 & 85.8 & 76.0 & 73.2 & 13.6 & 37.2 & 14.8 & 87.5 & 88.8 & 81.0 & 52.5 & 91.3 & 49.6 \\
Table-R1-Qwen2.5-7B                 & 30.6 & 87.6 & 79.8 & 78.1 & 19.8 & 43.1 & 20.0 & 83.7 & 88.0 & 76.2 & 93.0 & 96.4 & 70.8 \\
\addlinespace[0.3em]
\bottomrule
\end{tabular}
}
\caption{Ablation study results on model-level, task-level, and  formats.}
\end{table*}

\end{document}